\documentclass[10pt]{article}

\newcommand{\ourtitle}{%
This human study did not involve human subjects:\\
Validating LLM simulations as behavioral evidence 
}

\newcommand{\ourkeywords}{}

\usepackage{etoolbox}
\providetoggle{blind}
\providetoggle{showtodos}
\providetoggle{splitsections}
\settoggle{blind}{false}
\settoggle{splitsections}{true}
\usepackage{graphicx}
\usepackage{tabularx}
\usepackage{array}
\usepackage{longtable}

\usepackage{todonotes}

\newcommand{\jessica}[1]{\textcolor{red}{Jessica: #1}}

\providetoggle{neurips}
\providetoggle{icml}
\providetoggle{iclr}
\providetoggle{arxiv}
\settoggle{neurips}{false}
\settoggle{icml}{false}
\settoggle{iclr}{false}
\settoggle{arxiv}{true}

\iftoggle{neurips}{
    \iftoggle{blind}{%
        \usepackage[nonatbib]{ext/neurips_2025}
    }{%
        \usepackage[nonatbib,final]{ext/neurips_2025}
    }
    \usepackage[square,comma,sort,numbers]{natbib}    \usepackage[font=footnotesize,labelfont=bf]{caption}
    \usepackage{algorithm,algpseudocode}
    \newcommand{\authorinfo}[4]{{#1}\\{#3}}
    \title{\ourtitle{}}
    \author{%
        \authorinfo{Jessica Hullman}%
        {Computer Science}%
        {Northwestern University}%
        {jhullman@northwestern.edu}%
    \And
        \authorinfo{}%
        {}%
        {}%
        {}%
    \And
        \authorinfo{}%
        {}%
        {}%
        {}
    }
}

\iftoggle{iclr}{
    \usepackage{ext/iclr2025,times}
    \usepackage{algorithm,algpseudocode}
    \nottoggle{blind}{\iclrfinalcopy}{}
    \title{\ourtitle{}}
    \author{%
    \name Jessica Hullman\\%
    \addr Northwestern University 
    \And %
    \name \\%
    \addr %
    }
}

\iftoggle{icml}{
    \usepackage[square,comma,sort,numbers]{natbib}    \usepackage[font=footnotesize,labelfont=bf]{caption}
    \usepackage{algorithm,algpseudocode}
    \usepackage[]{ext/icml2025}
}

\iftoggle{arxiv}{
    \usepackage{ext/jmlr2e_preprint}
    \usepackage[font=small,labelfont=bf]{caption}
    \usepackage{algorithm,algpseudocode}
    \jmlrheading{1}{2025}{}{}{}{Hullman, Broska, et al.}
    \ShortHeadings{\ourtitle}{Hullman, Broska, et al.}
    \firstpageno{1}
}{}

\usepackage[T1]{fontenc} 
\usepackage{environ,comment}
\usepackage{amsfonts,bbm,bm,courier,nicefrac,microtype}
\usepackage{amsmath,amsthm,amssymb}
\usepackage{array,booktabs,tabularx,multirow,colortbl,hhline}
\usepackage{eqnarray}
\usepackage[inline]{enumitem}
\usepackage{wrapfig}
\usepackage{xifthen}% provides \isempty test
\usepackage{thmtools}
\usepackage{xspace}
\usepackage{tikz}
\usepackage{listings}
\usepackage{caption}
\usepackage{subcaption}
\usepackage{seqsplit}
\usepackage[group-separator={,},group-minimum-digits={3}]{siunitx}
\renewcommand{\thmcontinues}[1]{}
\usepackage{inconsolata}

\usepackage{graphicx,xcolor,placeins}
\usepackage[font={footnotesize},labelfont={bf}]{caption}

\usepackage{hyperref}
\hypersetup{colorlinks=true, urlcolor=blue, linkcolor=blue, citecolor=blue, linkbordercolor=blue, pdfborderstyle={/S/U/W 1}}

\renewcommand{\arraystretch}{1.2}
\newcolumntype{H}{>{\setbox0=\hbox\bgroup}c<{\egroup}@{}}
\newcolumntype{R}[1]{>{\raggedright\arraybackslash}p{#1}}

\usepackage{natbib}
\setcitestyle{authoryear,round,sort}
\setitemize{leftmargin=1em,topsep=0.1em,parsep=0.5pt,}
\setlist[enumerate]{leftmargin=*, label= {\arabic*.}, itemsep=0.5em}

\usepackage[capitalise]{cleveref}
\newtheorem{theorem}{Theorem}

\newtheorem{example}{Example}

\usepackage{algorithm}

\algnewcommand{\alginput}[2]{\Statex{Input:~#1}\Comment{#2}}
\algnewcommand\algorithmicinput{\textbf{Input}}

\algnewcommand\algorithmicinitialize{\textbf{Initialize}}
\algnewcommand\algorithmicbigstep{\textbf{Step}}
\algnewcommand\INPUT{\item[\algorithmicinput]}
\algnewcommand\INITIALIZE{\item[\algorithmicinitialize]}
\algnewcommand{\STEP}[1]{\item[\algorithmicbigstep]{\textbf{#1}}}
\algnewcommand{\InputExplanation}[2][.6\linewidth]{\leavevmode\hfill\makebox[#1][r]{~{\footnotesize{#2}}}}
\algnewcommand{\InitializationExplanation}[2][.6\linewidth]{\leavevmode\hfill\makebox[#1][r]{~{\footnotesize{#2}}}}
\algnewcommand{\StateComment}[2]{\State{#1}\InputExplanation{#2}}
\algnewcommand{\alginitialize}[2]{\Statex{#1}\InitializationExplanation{#2}}
\algrenewcommand\algorithmiccomment[2][]{#1\hfill\textit{\scriptsize{#2}}}

\usepackage{needspace}
\usepackage[framemethod=TikZ]{mdframed}
\newtheoremstyle{break}% name
 {}%         Space above, empty = `usual value'
 {}%         Space below
 {\itshape}% Body font
 {}%         Indent amount (empty = no indent, \parindent = para indent)
 {\bfseries}% Thm head font
 {.}%        Punctuation after thm head
 {\newline}% Space after thm head: \newline = linebreak
 {}%         Thm head spec

\newtheoremstyle{framed-and-break}  % follow `plain` defaults but change HEADSPACE.
  {0em}   % ABOVESPACE
  {0em}   % BELOWSPACE
  {\itshape}  % BODYFONT
  {0pt}       % INDENT (empty value is the same as 0pt)
  {\bfseries} % HEADFONT
  {\newline}         % HEADPUNCT
  {0pt\needspace{2\baselineskip}}  % HEADSPACE. `plain` default: {5pt plus 1pt minus 1pt}
  {}          % CUSTOM-HEAD-SPEC

\newtheoremstyle{framed-no-break}  % follow `plain` defaults but change HEADSPACE.
  {0em}   % ABOVESPACE
  {0em}   % BELOWSPACE
  {\itshape}  % BODYFONT
  {0pt}       % INDENT (empty value is the same as 0pt)
  {\bfseries} % HEADFONT
  {. }         % HEADPUNCT
  {5pt plus 1pt minus 1pt}%
  {}

\definecolor{thm-color}{RGB}{206, 225, 255}
\definecolor{corollary-color}{RGB}{206, 225, 255}
\definecolor{lemma-color}{RGB}{206, 225, 255}
\definecolor{proposition-color}{RGB}{206, 225, 255}
\definecolor{proof-color}{gray}{0.95}

\definecolor{definition-color}{RGB}{248, 222, 216}
\definecolor{assumption-color}{gray}{0.95}
\definecolor{example-color}{gray}{0.95}
\definecolor{remark-color}{RGB}{248, 238, 216}

\newtheorem{boxexample}[theorem]{Example}

{%
\begin{mdframed}[%
backgroundcolor=example-color,
linewidth=1pt,
innerleftmargin=0.5em,
innerrightmargin=0.5em,
innertopmargin=0.5em,
innerbottommargin=0.5em,
skipabove=0.5em, 
skipbelow=0.5em%
]%
\begin{boxexample}%
}%
{%
\end{boxexample}%
\end{mdframed}%
}

\DeclareMathOperator*{\argmin}{arg\,min}

\newcommand{\expect}[2][]{\mathbb{E}\ifthenelse{\not\equal{}{#1}}{_{#1}}{}\![{\def\givenn{\middle|}#2}]}

\newcommand{\prob}[2][]{\text{\bf Pr}\ifthenelse{\not\equal{}{#1}}{_{#1}}{}\![{\def\givenn{\middle|}#2}]}

\newcommand{\humandgp}{f^*}

\newcommand{\mpre}{M_{pre}}

\newcommand{\scen}{s}
\newcommand{\scenset}{S}
\newcommand{\scenspace}{\mathcal{S}}
\newcommand{\yRV}{Y_{\scen}}

\newcommand{\covs}{\mathbf{W_s}}
\newcommand{\covsnoscen}{\mathbf{W}}

\newcommand{\hyp}{H}
\newcommand{\hypspace}{\mathcal{H}}
\newcommand{\theory}{T}

\newcommand{\regress}{\hat{f_{\theta}}}

\newcommand{\train}{t}

\newcommand{\str}{\sigma}

\newcommand{\llm}[2]{\hat{f}\!\left(#1;#2\right)}

\usepackage{ifthen}

\begin{document}

\iftoggle{neurips}{%
\maketitle%
}{}

\iftoggle{icml}{
    \twocolumn[
    \icmltitle{\ourtitle{}}
    \icmlsetsymbol{equalpi}{*}
    \begin{icmlauthorlist}
    \icmlauthor{Jessica Hullman}{nw}
    \icmlauthor{}{}
    \icmlauthor{}{}
    \end{icmlauthorlist}
    \icmlaffiliation{nw}{Northwestern University}
    \icmlaffiliation{}{}
    \icmlcorrespondingauthor{Jessica Hullman}{jhullman@northwestern.edu}
    \icmlkeywords{\ourkeywords{}}
    \vskip 0.42in
    ]
}{}

\iftoggle{arxiv}{
    \title{\ourtitle{}}
    \author{%
    \name Jessica Hullman \email jhullman@northwestern.edu\\ \addr Northwestern University
    \AND
      \name David Broska \email dbroska@stanford.edu \\ \addr Stanford University
    \AND
    \name Huaman Sun \email hmsun@u.northwestern.edu \\ \addr Northwestern University
    \AND
    \name Aaron Shaw \email aaronshaw@northwestern.edu \\ \addr Northwestern University
    }
    \maketitle
}{}

\begin{abstract}
% Statem 
%
A growing literature uses large language models (LLMs) as synthetic participants to generate cost-effective and nearly instantaneous responses in social science experiments. However, there is limited guidance on when such simulations support valid inference about human behavior. We contrast two strategies for obtaining valid estimates of causal effects and clarify the assumptions under which each is suitable for exploratory versus confirmatory research. Heuristic approaches seek to establish that simulated and observed human behavior are interchangeable through prompt engineering, model fine-tuning, and other repair strategies designed to reduce LLM-induced inaccuracies. While useful for many exploratory tasks, heuristic approaches lack the formal statistical guarantees typically required for confirmatory research. In contrast, statistical calibration combines auxiliary human data with statistical adjustments to account for discrepancies between observed and simulated responses. Under explicit assumptions, statistical calibration preserves validity and provides more precise estimates of causal effects at lower cost than experiments that rely solely on human participants. Yet the potential of both approaches depends on how well LLMs approximate the relevant populations. We consider what opportunities are overlooked when researchers focus myopically on substituting LLMs for human participants in a study. %Other opportunities for benefiting from generative AI can be overlooked   
\end{abstract}

\iftoggle{arxiv}{
\vspace{0.25em}
%\begin{keywords}\ourkeywords{Generative AI, behavioral science, AI surrogates, silicon subjects, digital twins, }\end{keywords}
{}}

\section{Introduction}

A growing literature positions AI, and especially large language models (LLMs), as a transformative technology for simulating human behavior in 
%everyday interactions \citep{park2024generative}, professional roles \citep{liang2024}, and historical settings \citep{kozlowski2025simulating, kim2024aiaugmentedsurveysleveraginglarge}. Yet perhaps the most prominent proposal is to leverage the statistical regularities these models have learned to simulate responses in 
surveys or experiments. 
One common approach is to prompt an LLM with a set of background characteristics (e.g., age, gender, education), present the model with a description of an experimental condition, and ask it to predict how this type of person would respond if they participated in the study. A related strategy is to prompt the model with an archive of a person's prior survey responses, interview transcripts, and other personal information to construct a ``digital twin'' that predicts how that specific individual would respond \citep{toubia2025database, park2024generative}.

%Taken together, LLMs could support a more cumulative behavioral science in a field long characterized by low consensus \citep{almaatouq2024beyond, watts_should_2017, collins_why_1994}. 
%LLMs could support exploratory research by letting researchers probe and experimentally manipulate internal model representations (e.g., network activations). 
%Much of the excitement around LLMs for behavioral science is driven by the promise to serve as general-purpose human response interpolators: 
%that is, systems that can infer how a person would respond to any question and in any setting.
%models that can be asked to imagine how any described person would answer any question. 
%~\citep{messeri2024artificial}. %Rather than collecting domain-specific data and fitting a supervised model to predict a narrow, well-specified outcome, researchers can now supply an arbitrary textual description of a participant---demographics, motives, situational cues, or an assigned experimental treatment---and prompt the model to generate the participant’s likely response. 
%They reduce the high cost of running human subjects studies, potentially addressing problems of low statistical power in many published studies~\citep{button2013power,ioannidis2017power,sedlmeier1992studies}
 %paves the way .  
%By having multiple LLM agents interact, studies of human social behavior are similarly possible. 
If accurate, these ``AI surrogates'' or ``silicon samples'' promise to reduce persistent barriers in the behavioral sciences by lowering the cost and increasing the scale of studying human behavior. 
This promise sparks a debate over how best to leverage this new data source. While a growing number of empirical studies report that LLM-simulated responses can approximate patterns found in real survey and experimental data ~\citep{argyle2023out,hewitt2024predicting,cui2025large,binz_using_2023,dillion2023replace,li2024perceptual}, there is little consensus on what it means to show that AI surrogates are \textit{valid} for the purpose of studying human behavior. Here, we distinguish three strategies for validating the results of simulated studies and discuss their promises and limitations.
%.  Broadly stated, validity concerns whether estimates based on using AI surrogates as a data source can be expected to approximate, at least on average, the target parameters of human behavior in a survey, experiment or other setting. 

One line of thought, which we refer to as \textit{heuristic}, pursues a \textit{validate-then-simulate} approach~\citep{kozlowski2025simulating} that treats human and LLM-simulated subjects as interchangeable given evidence that LLM responses are a reasonable proxy for human responses in a related setting. For example, authors argue that LLMs' ability to reproduce the direction and significance of main effects for experiments in which participants play trust games means they can be used for studying human trust more broadly~\citep{xie2024can}, and that realistic willingness-to-pay and demand curves in their responses to price changes makes them a viable substitute for learning about consumer preferences~\citep{brand2023using}. 
More ambitiously, scholars have argued that LLMs can expand what is feasible with human subjects research: facilitating the collection of data on elites, small minorities, and other hard-to-reach groups \citep{jackson2025aibehavioralscience}; fielding longer questionnaires because simulated respondents do not fatigue \citep{dillion2023replace}; %testing research designs that would pose excessive ethical risk to human subjects \citep{bail_can_2024, grossmann2023ai}, running studies that would otherwise require extraordinary amounts of data, 
or simulating subjects from the past when there is no data \citep{kozlowski2025simulating,kim2024aiaugmentedsurveysleveraginglarge}.

Others propose a \textit{simulate-then-validate} approach, where AI surrogates are used in exploratory research to inform what study to conduct next. Here, the point is not to treat simulated results as definitive evidence, but to use them to prioritize which hypotheses and designs are most worth testing with human subjects. For example, researchers could simulate planned studies to surface issues with the design (e.g., the misunderstanding of a task), offer an indication of the direction of a treatment effect, and help select the most promising design from many possible options. 
%Moreover, simulation studies could help identify which findings may not replicate in a follow-up study, given that social phenomena can change over time, false-positive results can occur, and methodological limitations can undermine confidence in the original result. While the scientific literature is vast, resource limitations have led to replication studies being reserved for a very small proportion of published results. LLMs make it easy for behavioral scientists to recreate the original experimental conditions and outcome measures in a simulation study, check whether the reported effect reproduces under reasonable assumptions, and then prioritize replications with human samples for findings that appear fragile or inconsistent.

A separate line of work proposes \textit{statistical calibration} (e.g.,~\citealp{broska2025mixed}). This approach combines a sample of human participants with a typically larger sample of LLM-predicted responses. Under explicit assumptions, statistical methods prevent LLMs from introducing bias into point estimates of averages, regression coefficients, and other parameters. By augmenting the human sample with LLM predictions, statistical calibration offers more precise estimates at a lower cost than studies that rely exclusively on human subjects, though the precision advantages depend on the bias of LLM predictions. These gains in precision could help researchers conduct adequately powered hypothesis tests even when the required sample size would be prohibitively expensive. This includes studies targeting the small effects typical for the behavioral sciences~\citep{rauf_audit_2025}, interaction effects~\citep{gelman_you_2018}, or studies with many experimental conditions (e.g., \citealp{almaatouq2024beyond, voelkel2024megastudy}).

This diversity of approaches leaves individual researchers to grapple with questions that have high stakes for the future of behavioral science writ broadly: When can LLM outputs be treated as interchangeable with human data? When must they be statistically adjusted? What are the trade-offs of using them as exploratory tools? %While some critics have advocated loudly against the adoption of silicon subjects on ethical or fundamentalist grounds~\citep{}, 
% \jessica{I'm not sure what we are saying about why they are different here - my sense is that we could talk about this in Discussion, since it only raises questions for me as a reader here} %Hype directed for and against the use of LLMs in social science risks overshadowing 
In this paper, we analyze the forms of validation that underwrite current uses of LLM simulations in behavioral science and clarify the kinds of scientific claims each approach can responsibly support.
First, we characterize heuristic validation practices, providing an overview of the types of partial evidence researchers are citing as support for treating LLM and human samples as practically equivalent. We describe threats to interchangeability like evidence of systematic bias and the potential for memorization to drive results. 
%cataloging \textit{ex-ante} repair operations commonly deployed to bolster this appearance.
Second, we synthesize takeaways from emerging literature on what is needed for valid identification of target human parameters from LLM measurements~\citep{ludwig2025large,perkowski2025when} to argue that heuristic validation cannot, even under optimistic conditions, guarantee the absence of systematic biases in LLM responses. This makes heuristic substitution unsuitable for providing the kinds of trustworthiness guarantees behavioral scientists are accustomed to expect from confirmatory research. %We discuss recent proposals that avoid these challenges by instead settling for valid estimates of prediction accuracy for known task populations or constraining use of silicon samples to exploratory piloting. 
Third, we show that calibration-based approaches lay a more suitable epistemic foundation by making assumptions explicit and, when those assumptions hold, delivering unbiased estimates. 
Calibration-based approaches are not, however, a panacea: like other empirical tools, these statistical adjustments rely on conditions that researchers need to carefully justify, and may offer modest gains in light of highly variable behavioral outcomes. 
Fourth, we discuss ways in which the prevailing discourse on generative AI for behavioral science may overlook opportunities for better LLM simulations to improve theory and design analysis. These opportunities raise questions about how to evaluate exploratory and design based applications of LLMs such as hypothesis generation
\section{Behavioral Research Context}
\label{sec:prelim}

%We define the class of LLM-aided behavioral research studies that our work addresses. 

%\subsection{Behavioral research setting} 
%\label{sec:setup}
We assume a quantitative empirical behavioral research setting, in which a researcher hopes to draw inferences about a human data-generating process, $\humandgp$ by conducting a controlled study. 
%The study is associated with a set of data collection assumptions and procedures, including assumptions designating a target population on which $\humandgp$ is defined and a method of sampling from it. 
Conducting the study entails 
probing $\humandgp$ with some set of scenarios $\scen \in \scenspace$ representing the experimental prompt or instrument text shown to participants, and produces a dataset $D = \{(X_i,Y_i)\}_{i=1}^n$, where instances are assumed to be independent and identically distributed. 
Each observation consists of a vector $X_i=(S_i,A_i,W_i)$, %(with realizations $x=(s,w)$), 
where $S_i$ is the specific experimental or survey prompt text shown to participants, $A_i$ is a vector of treatment indicators, and $W_i$ is a vector of subject-level observables such as demographic information.  %Let $Y$ denote the human response generated by the gold-standard human process $\humandgp$. 
\subsection{Hypothesis-Driven Science}
We are concerned with hypothesis-driven behavioral science, where the goal is to assess how much support a human dataset $D$ provides for a hypothesis $\hyp$, 
%A hypothesis captures 
an empirical regularity that is thought to hold in the population on which $\humandgp$ is defined. 
%. Estimands are functionals of \(V\) (e.g., \(\mathbb{E}[V(X)]\), regression coefficients from moments \(g\big(V(X),X;\theta\big)=0\), average treatment effects \( \mathbb{E}\!\left[V(X\mid T=1)-V(X\mid T=0)\right]\), etc.). 
For example, in a between-subjects study comparing responses to two scenarios distinguished by prompts $\scen_1$ and $\scen_2$, a hypothesis might concern the ordering of the means:
\[
H:\; \mathbb{E}[Y|S=\scen_1] > \mathbb{E}[Y|S=\scen_2].
\]
The researcher applies a statistical procedure to produce estimates with uncertainty that capture what has been learned about $\humandgp$ with respect to the hypothesis. This usually entails applying a decision rule that maps from the sample space of results %  $\mpost$ dictates a decision rule $\resultfun: \mathcal{O} \to \resultspace$ 
 (e.g., parameter estimates, test statistics) to a finite decision space  (e.g., significant or not).
It is also typically accompanied by statistical guarantees about the relationship between an estimate $\hat \theta$ and a target parameter $\theta^*$ (e.g., a population mean).
%(e.g., enacted using a specific statistical test), mapping from the sample space of results $\mathcal{O}$ (e.g., test statistics) to a finite decision space $\resultspace$ consisting of $q$ possible results, $\resultspace = \{\resultfun_1, \resultfun_2, \ldots, \resultfun_q\}$ (e.g., significant or not significant given some chosen $\alpha$). 
%For example, for the simple between subjects experiment, $\mpost$ might involve calculating the sample mean in each condition $\hat{\theta}_{\scen} = \frac{1}{n_{\scen}}\sum_{n=1}^{n_{\scen}}\yRVi$ then testing the hypothesis $\hat{\theta}_{\scen_{a}} > \hat{\theta}_{\scen_{b}}$ by evaluating the difference $\hat{\theta}_{\scen_{a}} - \hat{\theta}_{\scen_{b}}$ relative to its sampling variability (e.g., using an unpaired two sample t-test). Here, $q=2$ and the output space is over test statistics, $\mathcal{O}=\mathbb{R}$, and $\resultspace=\{null, alternative\})$. %We use $\resultfun(o)$ to denote the realized result (e.g., $\resultfun(o) = \{null\}$).  % or consist of a set of $q$ bins for a continuous parameter estimate, or a set of different explanatory models evaluated against the data in model selection, and so on. 
%For a given sample $o \in \mathcal{O}$, the realized result is $\resultfun(o) \in \resultspace$.

%We distinguish empirical research pursuits by their epistemological goals. 
At an empirical level, realized results %$\resultfun_a(o)$, $\resultfun_b(o)$, and so on 
are assessed in light of hypotheses. %$\hyp_a$, $\hyp_b$, and so on. 
At a theoretical level, hypotheses %$\hyp_a$, $\hyp_b$, and so on 
are assessed in light of their support for a theory, %$\theory$, 
a set of propositions about latent processes or relationships related to human behavior. 
Behavioral research may be \textit{confirmatory}, where the researcher begins with a theory, which is thought to logically imply a hypothesis, and conducts an experiment to test if the hypothesis holds. %A number of assumptions and entailments underpin confirmatory research, which prior work has elaborated in greater depth (Oberauer and Lewandowsky YYYY). We do not recapitulate these here, except to underscore that they are rarely stated or met, and that the machinery of confirmatory hypothesis testing is thus often an inappropriate analytical framework. 
In contrast, in \textit{discovery-oriented} (theory-building) research, theories do not strongly imply hypotheses. Instead, a larger space of hypotheses %$\hypspace_{\theory}$ 
representing possible empirical regularities implied by a theory is searched to discover effects that would support it~\citep{oberauer2019addressing}. 
%In practice, behavioral research is often presented as confirmatory when it may be better considered exploratory due, for example, to researcher degrees of freedom~\citep{gelman2013garden,oberauer2019addressing}, but this issue is orthogonal to our goals.

\begin{comment}
\subsubsection{Generalization}
In theory, the results of an idealized randomized controlled experiment or survey collection are expected to replicate\footnote{Expected replication is subject to sampling error and theoretical constraints on the reproducibility rate of the result of interest, conditional on the elements of the original experiment \citep[see, e.g.,][]{buzbas2023logical}.} in new samples drawn under the population assumptions and sampling frame and subject to the same background knowledge and measurement procedures %($\mpre, \mpost$), 
or meaning preserving variations~\citep{buzbas2023logical}. 
Standard assumptions for valid inference of population effects include successful randomization of treatment assignment, a target uncertain outcome on a well-defined set, and a modeling strategy that appropriately accounts for random effects and residual covariate imbalance. 
\end{comment}

%While metascientists have criticized how the vagueness of many social science theories calls in question their testability~\citep{eronen2021theory,meehl1990summaries,oberauer2019addressing}, we do not restrict our discussion only to behavioral studies associated with ``strong'' theory.  

\subsection{LLM simulations}
\label{sec:llm_setup}
We are concerned with settings where the researcher's dataset includes some portion of LLM responses in place of measurements generated by $\humandgp$. 
An LLM $\hat{f}$ %$\llm{.}{\train}$ 
can be thought of as a blackbox text generator that takes in a text string
%$\str$ 
from some alphabet 
%$\alphabet$ 
and outputs a text response. 
An LLM is trained on some particular collection of strings (i.e., training data).
%, with $\train$ denoting the collection of strings it was trained on. 
%In contrast to the data collection methods $\mpre$ used in the human study, $\llmpre$ describes the LLM data collection procedure. 
%We use $\hat{Y}_{X}=\llm{X}{\train}$ to denote the random variable for the LLM’s response when prompted to produce a label corresponding to covariates $X$.

In some contexts we consider, both human and LLM responses are collected for the same instances, creating $D_{\text{shared}} = \{(X_i, Y_i,\hat{f}(X_i))\}_{i=1}^{n}$. 
Additionally, we assume the researcher is interested in making use of a larger collection of instances $D_{\text{LLM}} = \{(\tilde{X}_i,\hat{f}(\tilde{X}_i))\}_{i=1}^{N}$ in generating or testing hypotheses.

\begin{figure}
    \centering
    \includegraphics[width=0.75\linewidth]{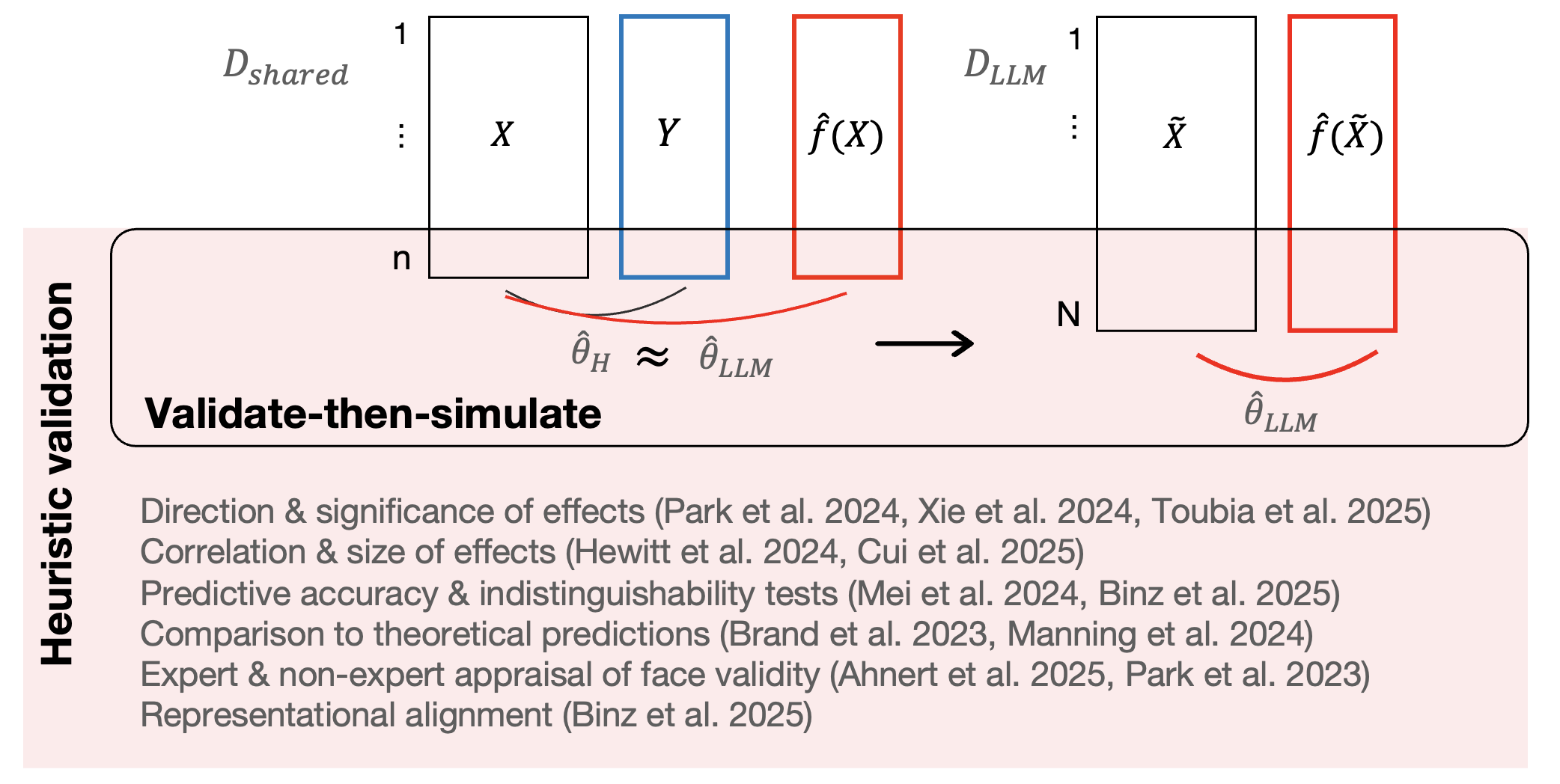}
    \caption{A validate-then-simulate approach involves collecting some human observations, which are jointly labeled by the LLM ($D_{\text{shared}}$), and using this dataset to demonstrate that the model achieves sufficient fidelity in approximating important aspects of the human results. Heuristic approaches then imply or assert that estimates ($\hat{\theta}_{LLM}$) derived from a larger dataset where only LLM labels are available ($D_{\text{LLM}}$) may serve as proxies for human-derived estimates ($\hat{\theta}_H$) without statistical adjustment.}
    \label{fig:validate-then-simulate}
\end{figure}

\section{Heuristic validation: From partial evidence of fidelity to generalization}
%\section{The heuristic path: From partial validation to generalization}
\label{sec:heuristic}

Heuristic approaches rely on evidence of observed alignment between LLM and human results in one setting to motivate treating them as interchangeable in a related setting where human ground truth is unknown. In a \textit{validate-then-simulate} approach~\citep{kozlowski2025simulating}, a jointly labeled dataset $D_{\text{shared}}$ is used to assess the extent to which LLM-based results align with human results, and conclusions consequently drawn about the fitness of the LLM for approximating related human behavior. 
For example, the researcher might confirm that the LLM responses provide similar support to human responses for some subset of hypotheses of interest. 
%$\mathcal{H}_{heur} \in \mathcal{H}$ of hypotheses deemed relevant to $\humandgp$. 
The LLM is then applied or presumed to be applicable to related scenarios where no human data is available (Figure \ref{fig:validate-then-simulate}). %, without explicit specification of how the new space of inputs compares to that on which human validation was presented. 
%, where the inputs $ are \textit{not} necessarily drawn from the same distribution. 
%Heuristic approaches thus involve the collection of LLM responses $D_{\text{LLM}} = \{(X_i,\llm{X_i}{\train}\}_{i=1}^m$, which are used to generate LLM results $\resultfun_{LLM}(o)$ that are then compared to human results of interest. 
 For our purposes, distinctions between whether authors explicitly assert or merely imply valid generalization of LLM simulations do not matter. Heuristic approaches do \textit{not} purport to offer statistical guarantees on the LLMs' generalization to the tasks of interest, relying instead on an implicit, partial validation approach.

The scope of the generalization target varies considerably across demonstrations of the heuristic approach. Heuristic validation has been implied to be sufficient for supporting "near generalization" of LLMs to novel scenarios where specific details (e.g., the number of rounds a game is played, or the number of agents on the board) are varied relative to scenarios validated against human judgment~\citep{tranchero2024theorizing}. Others apply the validate-then-simulate approach on a sample of scenarios to make broader claims, such as that an LLM can ``predict and simulate human behaviour in any experiment expressible in natural language'' \citep{binz2025foundation}. \citet{kozlowski2025simulating} suggest that evidence of interchangeability between human and LLM subjects in one context may be sufficient for using LLMs to study settings where human ground truth is difficult or impossible to attain. For example, researchers might apply LLMs to study hard-to-reach populations or questionnaires that are too long or complex to navigate for most humans, to obtain larger sample sizes than within their research budget, or to study settings that depend on unobservable counterfactuals, like attitudes in the past or future or new levels of ethical risk in well-known designs like the Stanford Prison experiment.

\subsection{``Validate-then-simulate'': Demonstrating LLM-human interchangeability}

Demonstrations of heuristic validation often take the form of replication-style experiments involving some gold standard human data and AI surrogates for comparison.
% where the results of LLM simulations are compared to human results at the level of individual experiments~\citep{aher2022using,cui2025large,hewitt2024predicting,xie2024can} or are assessed in light of human patterns distilled from sets of experiments~\citep{chen2025manager}. 
%In other cases, expectations of human behavior are more loosely defined based on prior research in an area~\citep{piao2025emergence}, or are dictated by theoretical models~\citep{chen2023emergence, horton2023large,manning2024automated}. 
We characterize several common heuristic evaluation patterns, drawing on eligible examples from \citet{anthis2025llm}'s sample of 53 empirical studies that compare human research subjects to LLMs as well as several more recently published demonstrations. Many papers in this sample simultaneously apply several of the evaluation patterns we describe below. They vary in the degree to which they assert the validity or generalization of LLM simulated data. We report more comprehensive descriptions of their validation methods in the Appendix.

\vspace{2mm}
\noindent \textbf{Direction and significance of hypothesized effects}. A coarse form of validation involves comparing the direction and/or significance of a subset of effects that are deemed behaviorally important across humans and LLMs~\citep{chen2025manager,lampinen2024language,van2025experimenting}.
In the largest example of this approach, \citet{cui2025large} find that LLMs replicate the direction and significance of up to 81\% of main effects reported in 156 randomly selected original studies~\citep{van2025experimenting}. However, LLMs also produce significant results for up to 83\% of effects that were not significant with human samples, depending on the model.
\citet{chen2025manager} examine the extent to which GPT models replicate the direction and significance of previously confirmed hypotheses about human biases, like the presence of negative autocorrelation in human attempts at generating random sequences (hot hand fallacy), while \citet{xie2024can} hypothesize patterns in human trust game behavior and confirm that these patterns are replicated by GPT models, and \citet{lampinen2024language} look for directional similarities in human and LLM performance on reasoning tasks. 
\citet{park2024generative} and \citet{toubia2025database} start with a set of social science studies, and compare the number of successful replications of their effects when 1000 and 2058 people they recruited, respectively, complete the studies, versus when generative agent simulations (or ``digital twins'') of those people complete the studies.  

%\vspace{2mm}
%\noindent \textbf{Study-level effect size and correlation}.
\noindent \textbf{Correlation and size of hypothesized effects}. Authors also examine correlations between study-level human and AI surrogate effect sizes. In addition to \citet{cui2025large}'s study of 156 experiments in psychology and management,  \citet{hewitt2024predicting} compare LLM-estimated effects for 70 preregistered social science survey experiments from an NSF initiative. Both sets of authors find moderate to high correlations (roughly 0.85 and 0.5 across the respective sets of studies), but that LLMs overestimate human effect sizes. \citet{ahnert2025extracting} find a wide range of correlation magnitudes between human survey and LLM responses to time-varying emotion and attitude survey items.   

\vspace{2mm}
\noindent \textbf{Predictive accuracy and indistinguishability tests}.
Some papers use predictive performance to validate LLMs that are prompted or fine-tuned to simulate behavior in surveys or behavioral experiments~\citep{park2024generative,binz2025foundation,kolluri2025finetuning,toubia2025database}. For example, \citet{binz2025foundation} validate Centaur, a model fine-tuned on psychology experiment data, by examining negative log likelihood averaged over responses when the model is used to predict responses for held-out participants (in-sample predictive performance) as well as held-out experiments (out-of-sample performance). 
\citet{park2024generative} use in-depth interview data from 1,000 people to construct LLM agent simulations, then examine individual-level accuracy, correlation, and mean absolute error in attempts to replicate individual responses to several surveys and controlled experiments. Using a measure of accuracy normalized by the individual's ability to replicate their original responses two weeks later, they find that the agent simulations are 85\% as accurate as the individuals themselves at replicating responses to the General Social Survey.  
Using a similar approach, \citet{toubia2025database} find that individual predictions from agents recover 88\% of the individuals' ability to self-replicate their responses to behavioral economics experiments.  
%\citet{kolluri2025finetuning} predict the individual response accuracy of a model fine-tuned on social science surveys and experiments, normalizing by the range of observed responses for the specific condition.  
%On the other hand, \citet{binz2025foundation} also report how poorly the Centaur model performs at predicting \textit{non-human} data designed to match aggregate statistics of human data as further support of the model's predictive validity.

%\vspace{2mm}
%\noindent \textbf{Distributional comparisons and statistical indistinguishability}.
Others argue for comparing human versus LLM response distributions directly, rather than comparing downstream estimates~\citep{kolluri2025finetuning}. Studies employ Wasserstein distance \citep{moon2024virtual}, KL divergence \citep{manning2025general}, total variation \citep{meister2024benchmarking}, and/or Earth mover distance (EMD) \citep{boelaert2025} to describe distributional fidelity. %Tests of statistical indistinguishability between LLM and human-generated distributions vary, and include parametric as well as non-parametric tests. 
Others, like \citet{binz2025foundation}, also examine the distribution of trial-by-trial response trajectories when each successive generation is conditioned on the model's own prior responses, based on an argument that this kind of ``model falsification'' approach~\citep{palminteri2017importance} provides a better test of generative ability.

Another common approach elicits non-expert human and/or automated evaluations of the fidelity of LLM simulations to human baselines.
Some authors use a Turing Test methodology~\citep{mei2024turing,wang2025experimental} designed to assess how difficult it is to distinguish LLM from human responses. Pairs of observations are repeatedly sampled from the LLM and human response distributions and compared to see which is more likely under the human distribution.% Authors then report the probability that the LLM response was more likely under the human distribution.
~\citet{park2023generative} instead elicit believability rankings from human evaluators to assess generative agent interaction transcripts against alternative agent architectures and human-produced baselines. \citet{hamalaninen2023evaluating} elicit both pairwise comparisons and open-ended written evaluations from non-experts to assess whether and how they discern human or LLM-generated narratives. Similarly, \citet{argyle2023out} present human raters with human and LLM-generated texts and ask them to evaluate contents and discern the origins of each. Such analyses are used to emphasize that the failure of human raters to discern the origins of data generated by people versus that generated by LLMs. Such failures purportedly provide evidence of the simulations' validity.

%While the above comparisons focus on generated response distributions for individual items or averaged over items,

\vspace{2mm}
\noindent \textbf{Comparison to theoretical predictions}.
Researchers have also compared LLM responses in economically-oriented tasks to theoretical predictions, interpreting consistency with theoretical predictions as evidence of a good simulacrum~\citep{brand2023using,chen2023emergence,manning2024automated,tranchero2024theorizing,akata2025playing,del2025can}. 
Frequently these comparisons suggest that LLMs better conform to rationality and other theoretical predictions in decision tasks~\citep{chen2023emergence,akata2025playing,manning2024automated}. %study GPT behavior in decision tasks that involve allocation under a fixed budget to examine the consistency of GPT decisions with a necessary and sufficient condition for utility maximization in a decision task. They find that relative to a human sample, GPT performs more rationally across tasks and demographics, though its rationality decreased significantly upon contextual changes like the prompt using a less standard presentation of prices. Similarly, \citet{akata2025playing} describe the rational but non-human-like ``reluctance to forgive'' of GPT 4 when put in a classic prisoner dilemma scenario, where it never cooperates again when playing an agent who defects initially, regardless of their subsequent behavior.  
Alternatively, \citet{tranchero2024theorizing} compare LLM simulation results and human results to an optimum defined in a mathematical formalization of a ``streetlight effect'' observed in human search behavior, where people tend to search near existing data.  
%\citet{manning2024automated} simulate an open ascending price art auction among bidders with private values, finding a pattern predicted by auction theory in which increasing  bidders' reserve prices increases the clearing price to close to the second-highest reserve price. %\citet{del2025can} compare the degree to which LLMs exhibit rational expectations in a price prediction setting, finding that the specific model version influences whether GPT market dynamics converge slower and faster than those observed with humans. 
Similarly, while \citet{brand2023using} affirm that when the price is varied for a good, GPT demand curves are downward sloping as predicted by economic theory, they also consider it promising that GPT demand curves are not \textit{monotonically} downward-sloping, arguing that it allows room to discover new phenomena characterizing human behavior. % captured by the model. 

\vspace{2mm}
\noindent \textbf{Expert appraisal of face validity}.
Authors sometimes report on how LLM results align qualitatively with expectations based on domain expertise. %Here, LLM results are subject to an implicit test $\pi$ dictated by the expert's expectations of human data: $\pi(\resultfun_{LLM}(x))$. 
For example, \citet{argyle2023out} comment on the face validity of GPT-3 responses based on qualitatively similar sets and frequencies of words from humans in a political context, while \citet{ahnert2025extracting} eyeball whether time series of LLM-imputed survey responses capture important trends in public emotions throughout the covid-19 response.

Some authors advocate for using the elicited reasoning from an LLM subject as a means of corroborating its simulation fidelity. For example, \citet{tranchero2024theorizing} suggest using AI surrogates to explore mechanisms that explain observed effects, which is ``greatly aided by the ability to prompt AI agents to justify strategic choices.'' \citet{xie2024can} appraise how rationales elicited from LLMs convey beliefs, desires, and intentions consistent with their decisions in behavioral economic trust games, using the rationales alongside quantitative results. % to conclude that ``humans’ trust behavior, one of the most elemental and critical behaviors in human interaction across society, can effectively be simulated by LLM agents.'' 
\citet{binz2025foundation} go a step further, testing the utility of an LLM's explanation by using it to instantiate a new cognitive model that implements the reported strategy, then examining the prediction accuracy of this new model. %, which predicted human responses better than strategies considered by the original study. 
%As we discuss below, there are risks to unquestioning reliance on behavioral explanations generated by LLMs.

\vspace{2mm}
\noindent \textbf{Representational alignment}.
Inspired by prior evidence of the predictive power of language models' internal representations for human neural activity~\citep[e.g.,][]{schrimpf2021neural,zhou2024divergences},
\citet{binz2025foundation} examine how well models' internal representations predict human neural activity as recorded in functional magnetic resonance imaging (fMRI) measurements. 
They use probing, where an outcome variable of interest---in this case fMRI measurements---is regressed on an LLM's internal activations. 
They argue that a consistent difference in correlations between model activations and human data for their fine-tuned model relative to an unmodified model %, including for out-of-distribution stimuli, 
suggests that fine-tuning leads the model to better generalize.

\subsection{Threats to heuristic validation}
The presence of systematic bias and error in LLM responses or human ground truth poses challenges to heuristic validation. We summarize several threats.

\subsubsection{Systematic bias and distortion in AI surrogates}

For heuristic validation to be trustworthy, we would expect to see high resemblance to human results with little evidence of significant biases or distortions contributed by LLM proxies. Yet evidence to the contrary is shown in multiple behavioral domains.

As mentioned above, large scale replication studies describe moderate to strong correlations between effect sizes in human and LLM results~\citep{hewitt2024predicting,cui2025large}, but effect magnitudes that were larger on average than those derived from humans. 
%This can result when 
LLM response distributions can fail to match critical statistical moments of human response distributions such as variance, quantiles, or skewness. 
The fact that multiple comparative studies have reported lower variance in LLM-produced response distributions compared to humans \citep{bisbee2024synthetic,kaiser2025simulating,murthy2024one,wang2025large,boelaert2025, park_diminished_2024} is problematic whenever the goal is to characterize uncertainty as well as magnitude. This affects, for example, estimates of standardized effects (Cohen's d, probability of superiority, etc.) or tests for significance, both of which have been central in human behavioral science.  %This might not be problematic when estimating mean responses for survey items or experimental conditions; however, behavioral science is frequently concerned with estimating standardized effects, where differences in variance change the magnitude of the estimate.  % where characterizing uncertainty is critical.
%Heuristic evaluation inherits challenges associated with replicating behavioral results more broadly, including reduction of comparisons to dichotomous statements about whether replication results were significant in the same direction, and heavy dependence of results on statistical power. 

Asymmetries have also been observed in the fidelity of reproduction across groups, with prediction accuracy and variance potentially being lower for subgroups such as historically disadvantaged groups, older populations, or other groups that are mis/underrepresented on the internet~\citep{durmus2023towards,dominguez2024questioning, lee2024large,wang2025experimental}. For example, authors have found LLMs to be significantly less likely to replicate studies involving ``socially sensitive topics'' like race and gender (e.g., a drop from 77\% to 42\% main effect replication rate~\citep{cui2025large}). %, %GPT-4's main effect replication rate, for example, dropped from 76.8\% in studies without race variables to 41.5\% in studies with race variables.
%and that LLMs do not reflect natural variations in entropy across questions among respondents from a subgroup~\citep{dominguez2024questioning}. %, and thus their alignment with subgroup response distributions can be predicted by how close that group's aggregate statistics are to a uniform distribution. 

Related results describe caricaturing of identity~\citep{cheng2023compost,liu2024evaluating,wang2025experimental}. For example, \citet{wang2025large} use embeddings to represent free-text responses to opinion-style questions, including what it is like to have a certain identity. Comparing the similarity of demographic identity-prompted LLM responses to human in-group and out-group representations, they find that the LLM responses tend to more closely resemble responses that out-group members expect of a group than what members of that group say themselves. %\citet{cheng2023compost} similarly observe exaggeration in LLM responses of the defining characteristics associated with a persona, particularly for nonbinary gender, non-white race/ethnicity, and political leaning, at the expense of characteristics associated with the topic of the question. 
\citet{liu2024evaluating} demonstrate that LLMs are less consistent when generating responses for incongruous personas--those where the view taken (e.g., an open attitude toward immigration) conflicts with that expected based on the persona profile (e.g., political conservative). 
Taking a broader perspective, \citet{crockett2025ai} argue that the generalization power of AI surrogates for social science is fundamentally limited by historical and structural conditions upon which AI tools are built, including the non-representative composition of model training datasets.

%\jessica{move somewhere:} A risk of relying heavily on heuristic checks of whether the direction and significance of human results is achieved by LLMs (as many authors are doing) is that they can miss important differences in effect magnitude and uncertainty. 

\subsubsection{Memorization}
If similarity between human and LLM responses is driven by the presence of prompt information (i.e., the experimental scenario text component $S$ of the input $X$) in the training data, this would make it harder to predict similarity for other related scenarios. Hence, when authors imply that valid LLM simulation of human behavior will generalize to scenarios they did not study, the default assumption is that neither the studied scenarios nor the novel related scenarios appear in the model's training data (a.k.a., ``leakage''). However, attempts to explicitly check for potential overlap between study prompts and model training data remain rare in heuristic validations, as do attempts to validate on brand new scenarios. % that could not be in the training data . 
In their recent review, \citet{anthis2025llm} observed that only one of $53$ studies---namely, \citep{hewitt2024predicting}---compared the predictions of LLMs on novel scenarios with human predictions of the same data.

\subsubsection{Misleading generalization} %}
Similarities between LLM responses and human judgments can lead researchers to expect LLM performance to generalize to related tasks in the same ways that human performance generalizes. 
%overinterpret LLM behavior despite the possibility that it was generated by fundamentally different mechanisms. 
%Because researchers are accustomed to interpreting human responses as evidence of underlying cognitive states, preferences, or strategies, they may fall back on interpretations that do not accurately describe LLM processes or capabilities.
%Model responses often display what \citet{anthis2025llm} call alienness: they appear human-like but are generated by processes with no psychological correspondence. 
%However, the landscape of plausible errors an LLM can make is often very different from those humans might make. For example, 
LLMs have been shown to demonstrate % can fail due to their autoregressive (predict-the-next token) training objective, demonstrating 
struggle on tasks involving compositionality, spatial or temporal reasoning, or minor prompt rephrasings~\citep{mccoy2023embers,tjuatja2024llms,kozlowski2025simulating}. Such brittleness undermines researchers' ability to infer that behavioral regularities reflect stable internal representations. 
For example, if an LLM can correctly define a topic (e.g., metaphor), we would also expect it to do well at classifying example texts as containing metaphors or not, but models frequently demonstrate ``potemkin understanding'' \citep{mancoridis2025potemkin} where they perform well on one test of a concept but poorly on another that humans who understood the concept would find easy. %Such failures directly challenge researchers' attempts to argue for generalization of LLM capabilities based on their human-like  performance on some subset of tasks. % as equivalent to human-style conceptual competence, ignoring that the space of possible misunderstandings for a human may be much smaller and more constrained than for an LLM. %Such patterns exemplify the ``alienness'' \citep{anthis2025llm} of model responses and cognition, which may appear human-like in many ways that contribute to ``illusions of understanding'' \citep{messeri2024artificial}. 

Researchers who are used to corroborating patterns in recorded behavior by querying subjects for their decision strategies or rationales must be cautious in doing the same with LLMs. %Human subjects can be probed for strategy use or reasoning, and their reported rationales---despite known limitations---interpreted within a common cognitive framework. 
A growing body of work suggests that the explanations of reasoning that LLMs provide are fragile~\citep{kambhampati2024can,lanham2023measuring,turpin2023language,paul2024making,madsen2024self} and driven by token associations~\citep{tang2023large}, such that their plausibility depends heavily on the relationship between the training data and test setting~\citep{zhao2025chain}. For example, chain-of-thought explanations can fail to mention aspects of the inputs that systematically affect their predictions~\citep{turpin2023language,arcuschin2025chain,chen2025reasoning}, and their faithfulness can vary considerably across tasks or models~\citep{lanham2023measuring}. While human responses cannot necessarily be trusted to be perfectly accurate and interpretable either, researchers have spent years developing intuitions about why human responses might be inaccurate as well as methods for detecting such inaccuracies. As simulations of human text, LLMs do not necessarily solve these problems, and introduce the need to develop different kinds of intuitions about and measurements of error as well. 
%suggests researchers cannot rely on the usual logic in which consistent behavior plus plausible rationales support an inference of understanding

\subsubsection{Noise, bias, and underspecification in human gold standard}
To the extent that the behavioral targets of studies are underspecified or open to interpretation (e.g., ``helpfulness'' in bystander studies), LLMs are of limited value. Most applications of LLMs to behavioral science experimentation to date are orthogonal to foundational methodological issues like the inversion of hypothesis testing to enable weak tests of theories relative to the hard sciences~\citep{meehl1990summaries}.

Additionally, evidence of low replication rates among well-cited empirical results in the behavioral sciences \citep{open2015estimating} suggests cautious interpretations whenever individual studies are held up as human ground truth to argue for sufficient AI surrogate fidelity. %The statistical power of a study design is the probability with which we expect to detect an effect of a given size when present under our study design. 
For example, methodologists have criticized the low statistical power observed in many behavioral studies \citep{sedlmeier1992studies,button2013power,ioannidis2017power}, which leads to a noisy published record. 
Few papers in our sample of heuristic validation examples engaged with the potential for bias and noise in the human ground truth to which they compared AI surrogates.\footnote{Exceptions are several papers that develop simulations from extensive data on specific individuals, where the authors first acknowledged where the human participants failed to replicate known experiment results~\citep{park2024generative,toubia2025database}. %These papers also acknowledged noise in the human ground truth by normalizing silicon sample prediction accuracy estimates by humans' test-retest accuracy when reprompted with the same questions. 
Another exception is \citet{cui2025large}, who interpret correlations between human and LLM results results in light of correlations between prior human replication experiments.} 
By not acknowledging the noisiness of individual study results, heuristic approaches to LLM validation risk further entrenching a view that small samples of human responses can be interpreted as reliable estimators of human behavior.

\subsection{Ex-ante repair strategies to improve predictive power}
In response to some of the challenges above, researchers propose \textit{ex-ante repair operations}: methods that optimize LLM data collection prior to estimation or inference to maximize human resemblance. While a comprehensive summary is beyond the scope of this paper, common ex-ante repair operations include advice on optimal model size~\citep{argyle2023out,binz2025foundation,hewitt2024predicting,zhu2025using}, configuration of hyperparameters like temperature~\citep{chen2023emergence,cui2025large,wang2025experimental}, prompt optimization~\citep{aher2022using,santurkar2023whose,dominguez2024questioning,park2024generative,toubia2025database}, distributional prompting~\citep{meister2024benchmarking}, ensemble prompting~\citep{hewitt2024predicting,xiedistributional,manning2025general}
response formatting and ordering~\citep{dominguez2024questioning,moore2024large,wang2025experimental,raman2025reasoning}, and supervised and reinforcement learning-based model fine-tuning~\citep{binz2025foundation,suh2025language,huang2024social}. 
Many heuristic validation attempts are accompanied by relative comparisons of LLM responses under ex-ante repair operations to simpler methods (e.g., simpler prompting strategies, or base models or existing cognitive models) to illustrate how researchers have improved the fidelity of the LLM simulations. % and reduce risks of bias. 
Such improvements in fidelity may be useful, as we demonstrate in our discussion of statistical calibration methods below. However, they do not provide a basis for valid inference.
\section{Necessary conditions for valid inference from LLM surrogates}
While ex-ante repair strategies aim to bring LLM results closer to human results, heuristic approaches rest on implicit arguments about sufficient closeness. %, rather than attempting to directly account for the bias that LLMs may contribute. 
%But behavioral researchers routinely accept that statistical methods come with conditions that must be met for their results to be credible--consider, for example, valid estimation of treatment effects in causal inference requiring conditional independence of treatment assignment and potential outcomes, or Ordinary Least Squares linear regression assuming mean zero errors that are uncorrelated with regressors and have equivalent variance. 
In this section, we summarize what conditions must hold for LLM responses to be treated interchangeably with human responses for the purposes of predicting and drawing inferences about human behavior. %We also discuss how relaxing the goal, from valid confirmatory plug-in estimates using LLMs to valid estimates of predictive accuracy for a known population of tasks, sidesteps thorny assumptions but may have limited value for inferential goals. 

%For example, few authors take explicit steps to check whether information about the study is included in the LLM's training data. However, we would expect overlap between the study materials and LLM's training data to result in underestimation of its prediction loss on new scenarios. 
%When only heuristic validation approaches of the types we describe above are used, there is no assurance that LLM-based results will generalize to new human samples under the same conditions , much less different conditions as authors imply. 
%following \citet{ludwig2025large} for why such biases threaten generalization from LLM estimators based on assuming interchangeability. 

\subsection{When is simple substitution of LLM responses sufficient?}
\label{sec:conditions}
To understand why heuristic validation cannot ensure valid generalization, it is worth asking under what conditions a researcher \textit{can} trust behavioral results obtained from LLM simulations for unseen tasks.  
Recent work by \citet{ludwig2025large} provides an econometric framework to clarify the conditions under which substituting a blackbox LLM’s responses for target human measurements provides a valid plug-in estimate for downstream parameter estimation. We summarize their results with minimum notation for accessibility, and present a more formal summary in the Appendix. We refer the reader to their manuscript for full details and proofs.
\vspace{2mm}

\subsubsection{Generalization target}  
One way to conceive of the \emph{generalization target}---the space of behavioral tasks or items for which the researcher wants to claim LLMs provide sufficient substitutes---is the population distribution over study scenarios (i.e., the prompt component $S$ of $X$) from which the researcher imagines the stimuli in their study are sampled. Consider arguments that experimental stimuli should be treated as samples rather than the population~\citep{clark1973language}: typically researchers want to generalize not only to new subjects facing the exact same stimuli or information their subjects faced, but new subjects experiencing stimuli or information of the same type. 
Consequently, beliefs about how experimental scenarios are sampled for study become important to what we can claim from our results. 

When the goal is to generalize from behavioral studies that involve AI surrogates, we must also account for beliefs about the sampling of scenario texts in the model training data. If, for example, the scenarios on which we heuristically validate LLM responses against human appear in the training data, but those we hope to generalize to do not, then we should not trust that the resemblance we observe in the former will hold to the same degree in the latter. 
Hence, the \textit{research context} a researcher intends to generalize to is defined by their beliefs about how the studied experimental scenarios and the training data are sampled. Imagine that the researcher were to repeatedly sample the same number of experimental scenarios on average from the population of relevant ones, where sampling is independent but not all scenarios are necessarily sampled with equal probability; i.e., the researcher does not necessarily study a random sample. What must we know about the generating process to treat behavioral results obtained by using LLMs interchangeably with those obtained from human subjects for the purposes of generalizing to a population of related scenarios?

\citet{ludwig2025large} show that two high level properties of a model's behavior are sufficient 
for an LLM that achieves them to be sufficient for studying human behavior in a research context.  
\vspace{2mm}

\noindent\textbf{Condition 1: No Training Leakage.}
\label{sec:cond_no_leakage} A first condition, No Training Leakage, is required for the researcher to extrapolate the predictive ability of an LLM on scenarios $D_{\text{shared}}$ to the larger population of relevant scenarios. To treat the LLM's predictive ability on the sampled scenarios as an unbiased estimate of its ability on the larger population, the expected sample average loss computed for the LLM's predictions should recover its average loss over the researcher's target collection of inputs. 
Imagine, however, that one or more text components in the set that the researcher sampled also appear in the LLM's training data. 
Then we would expect the sample average loss computed from the LLM's predictions to overstate its true performance on scenarios of interest. 
Thus, without placing further restrictions on how scenarios are sampled (such as requiring that the observed instances in $D_{\text{shared}}$ mirror the population mixture over the research context), for LLM use to yield unbiased estimates of the predictability of human responses, the LLM should not have been trained on any scenario in the population from which relevant scenarios are drawn. If there is reason to believe that some experimental prompts from the broader population of relevant scenarios have positive probability of appearing in the LLM training data, then we can't learn directly about the probability of certain human responses from using the LLM as a stand-in.

No Training Leakage is a concern whenever the goal is to generalize beyond the specific set of sampled scenarios. If the researcher observes all the relevant experimental prompts in the population of interest, then there is no concern, because there is no generalization. 

If extrapolating the LLM's predictive performance (or \textit{risk}) on a subset of scenarios where human data is available is the researcher's primary goal, one can get around the No Training Leakage requirement by 1) explicitly defining a finite population of relevant scenarios, such as by combinatorially varying task parameters (e.g., number of players, budget endowment in economic cooperation games), and 2) randomly sampling the scenarios for which joint labels are attained.\footnote{Here, scenarios must be independent and identically distributed, and the human responses should also be independently and identically distributed within each scenario.} \citet{manning2025general} demonstrate this, partitioning the scenarios in $D_{\text{shared}}$ into two non-overlapping samples, where one is then used to find the best combination of prompts to predict human behavior and the other used for estimating prediction accuracy. Under standard assumptions,\footnote{\citet{manning2025general} assume finite second moments of the loss to ensure asymptotic normality and positivity--i.e., that the LLM assigns nonzero probability to human outcomes--if the common log-loss measure is used.} this allows them to estimate the prediction accuracy of the LLM over the defined population. Because the population is known and the scenarios were randomly sampled, it is straightforward to calculate confidence intervals to characterize uncertainty in the estimate. Because hold-out testing is used, the estimator remains valid even if the LLM has memorized some of the sampled prompts or exhibits systematic bias in mimicking human responses.

\vspace{2mm}
\noindent\textbf{Condition 2: Preservation of Necessary Assumptions for Parameter Identification.} 
\label{sec:cond_no_error}
A second condition, Preservation of Necessary Assumptions for Parameter Identification, is required for downstream inferences about behavioral parameters to be valid under substitution of human responses with LLM responses.

Consider an idealized case, where we know that with high probability, for all scenarios we might sample in our study, the LLM's prediction of the human response has small error. 
We might think that in such a case, we could also expect any parameter estimates based on modeling the data (e.g., regressions to estimate causal effects of experimental manipulations) to closely approximate the estimates we'd get from using the same amount of human data. 
However, this is not the case. Even when prediction error is low, bias in parameter estimates in a downstream analysis can be large if LLM errors are correlated with covariates of interest.

As an example, suppose the goal of an analysis is to regress an outcome variable ($Y$)--say, donation amount--on a persona indicator--say, political affiliation ($Z$):
\[
Y \;=\; \beta_0 + \beta_1 Z + \eta,
\qquad Z\in\{0,1\} ,\quad \mathbb{E}[\eta\mid Z]=0
\]

Imagine the researcher instead runs the LLM equivalent:

\[
\hat Y \;=\; \tilde\beta_0 + \tilde\beta_1 Z + \eta_{LLM}
\]

Let $\varepsilon=\hat{Y}-Y$ be the LLM measurement error. Even if the expectation of this error is 0, if it varies systematically with $Z$ (e.g., the predictions are slightly over for $Z=1$ and slightly under for $Z=0$), then we should expect our estimate of the coefficient $\tilde\beta_1$ to be off from the true human estimate.\footnote{See the Appendix for an example showing that for a single regressor OLS scenario, the bias is proportional to $\frac{Cov(Z, \varepsilon)}{Var(Z)}$.} 
For example, in an empirical demonstration, 
\citet{egami2024using} use simulation to show that for a binary human label prediction task, LLM predicted labels that achieve $90\%$ prediction accuracy can, if plugged into a regression, produce an estimator with large bias (roughly 30\% relative to the unbiased estimator) and 95\% confidence intervals that cover the true value only about 40\% of the time.

What then must be true in order to treat LLM responses as interchangeable with human ones without risk of LLM-contributed bias? Behavioral researchers are accustomed to expecting that certain conditions must hold for valid estimation (\textit{identification}) of a parameter value like a mean or a regression coefficient. 
For example, researchers accept that to estimate a treatment effect using a randomized controlled experiment requires conditional independence of the treatment assignment and potential outcomes under each treatment. 
As a general rule, when LLMs are used instead of human subjects, at a high level, the mechanics of statistical inference that enable parameter identification must not be affected by the switch to AI surrogates.

Preserving assumptions for parameter identification means preserving \textit{moment conditions} on which inferential models depend. Many parameters that a researcher might target in a behavioral study---including means, treatment effects, or regression coefficients---are defined as the values that make certain population averages (or \textit{moments}) equal to zero. 
For example, assume the target parameter is a population mean $\mu$, and the true value of the mean is $\mu^*$. Let $V(x)$ be the expected value of the outcome for input $x$: $V(x)=\expect{Y|X=x}$, and $V(X)$ be the corresponding random variable. The moment $g(V(X),X;\mu)$ is the deviation $V(X)-\mu$ which satisfies $\mathbb{E}[V(X)-\mu]=0$ at the solution $\mu^*=\expect{V(X)}=\expect{Y}$.
If an LLM is to substitute for human measurements without distorting inference, it must preserve those moments: replacing human outcomes with LLM outputs should leave the relevant population averages that identify the parameter solution unchanged.

Consequently, \citet{ludwig2025large} show that in addition to No Training Leakage, a researcher would need to establish that for any parameter value, the expectation of the moment function using responses $\hat{V}(X)=\mathbb{E}[\hat{Y}|X]$ from an LLM is equal to the expectation of the moment function given human responses $V(X)$\footnote{In other words, $\mathbb{E}\!\big[g\big(\hat V(X),X;\theta\big)\big]\;=\;\mathbb{E}\!\big[g\big(V(X),X;\theta\big)\big]$, for all parameter values $\theta\in\Theta$, where expectations are taken over the distribution $P_X$ of analysis inputs. See the appendix for additional assumptions made by \citep{ludwig2025large} to derive this condition.}. \citet{perkowski2025when} presents related results in a causal inference framework (see the Appendix in \ref{sec:appendix}).

%\footnote{\citet{ludwig2025large} make several additional assumptions of the moment condition $g(\cdot)$ to derive these results: first, that the moment condition is differentiable in $\theta$ and the derivative uniformly bounded, and second that the moment condition is sensitive to $V_{x}$, by requiring a strictly positive lower bound on the partial derivative of $g(\cdot)$ with respect to the plugged-in measurement $v$ for all $v$ and $\theta$.}:

No training leakage alone is \textit{not} a sufficient condition to achieve valid parameter identification: even if the researcher were to know that the maximum error across any input $x \in \mathcal{X}$ with positive probability of being sampled were bounded, it is still possible that text generators may exist that satisfy these bounds, but exhibit (potentially small) errors that nonetheless non-trivially bias parameter estimation~\citep{ludwig2025large}.
The potential for the LLM’s errors to be correlated with covariates within the analysis input $X$ thus presents a pernicious problem that cannot be overcome by No Leakage alone.

Consequently, the researcher who wishes to blindly substitute LLM responses for human ones must first demonstrate no leakage between experimental prompts and the model training data. In other words, only models with documented training data should be used. The largest proprietary models, which researchers have observed best mimic human responses~\citep{argyle2023out, binz2025foundation, hewitt2024predicting, zhu2025using} typically do not release training data, providing only training data cut-offs.  %This is possible  but implies a trade-off in generation fidelity, given the many papers that observe higher fidelity human behavioral simulations from larger models. 
Thus the researcher must either train their own models, restrict their use to models with open training data \citep[e.g.,][]{biderman2023pythia}, or only study settings that they can verify post-date the model's training data cut-off.  
Second, the researcher must demonstrate that the model predictions do not affect their target moment conditions for any parameter value. 
This is much less feasible to achieve. After all, if the researcher could be confident this was the case, they would no longer need to rely on LLM simulations. 

\begin{figure}[ht]
    \centering
    \includegraphics[width=0.7\linewidth]{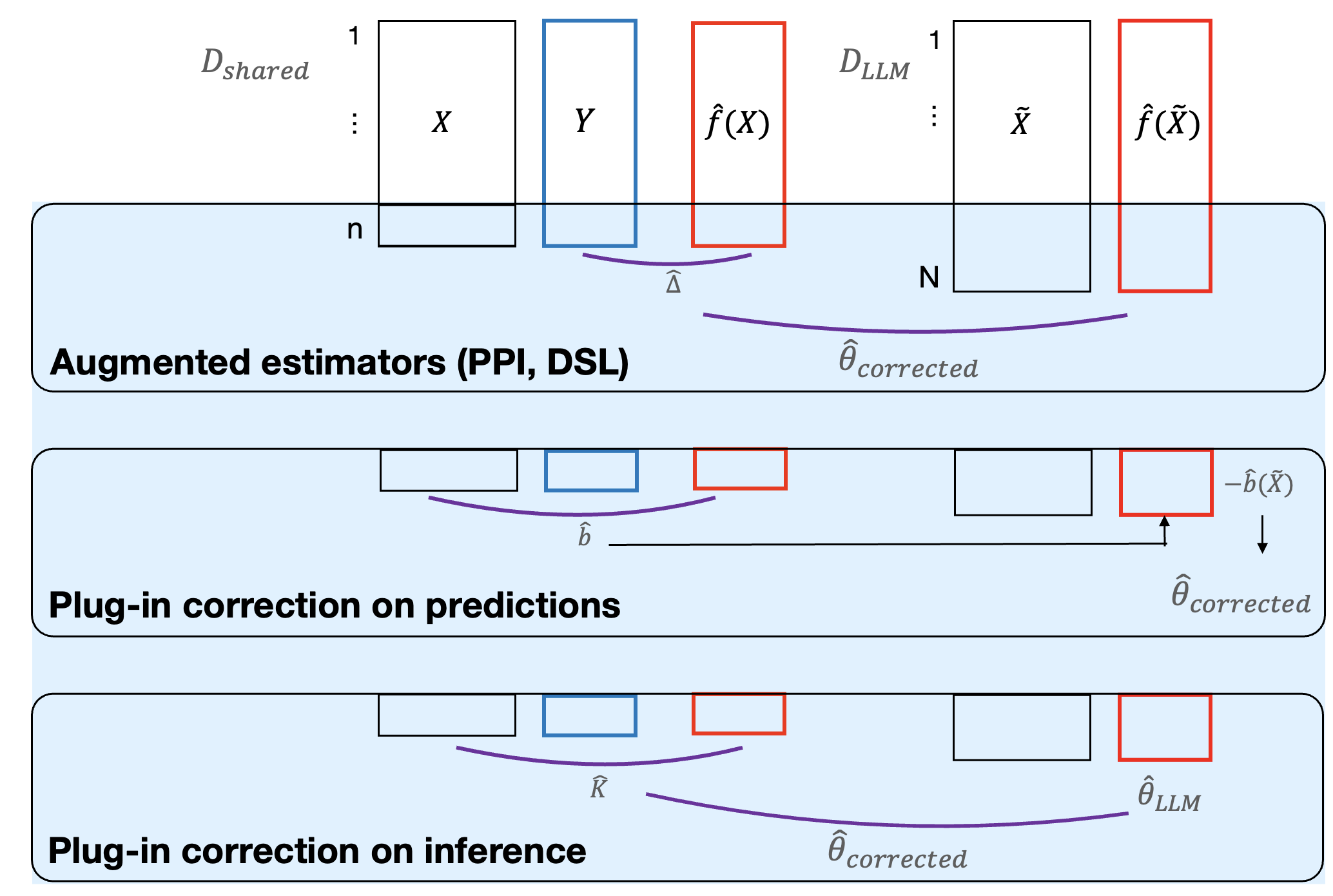}
    \caption{Statistical calibration approaches derive estimators that explicitly account for bias contributed by LLM approximations to human responses. PPI~\cite{angelopoulos2023prediction}, DSL~\cite{egami2024using}, and related approaches learn a rectifier and additively combine it with a base estimator. Approaches to plug-in correction learn a model of the relationship between human ground truth and LLM predictions using jointly labeled data ($D_{\text{shared}}$), then correct either the LLM predictions prior to estimation~\citep{ludwig2025large} or directly adjust the target inference~\citep{wang2020methods}. }
    \label{fig:calibration}
\end{figure}

\section{Statistical calibration: Explicit accounting for bias in inference}
\label{sec:address_error}
An emerging literature combines samples of human responses with LLM-predicted responses to estimate population means, regression coefficients, and other parameters \citep{wang2025marketllms,broska2025mixed,ludwig2025large}. In what we call the \textit{statistical calibration} approach, the LLM is treated as a cost-effective but imperfect source of information about human behavior. This approach uses statistical techniques to calibrate the LLM-predicted responses against the human sample to prevent the LLM from introducing bias into parameter estimates. Depending on how accurately the LLM predicts human responses, combining the two data sources can yield more precise estimates at a lower cost than solely relying on human data.

Calibration approaches assume a dataset $D_\text{LLM}$$=\{\tilde X_i, \hat{f}(\tilde X_i)\}^N_{i=1}$ containing inputs $\tilde X_i$ (e.g., scenario text and respondent features) and their predicted outcomes $\hat{f}(\tilde X_i)$, as well as a typically much smaller dataset $D_\text{shared}=$ $\{(X_i,$ $Y_i,$ $f(X_i))\}^n_{i=1}$, which includes inputs $X_i$, their observed human outcome $Y_i$, and LLM-predicted outcome $\hat{f}(X_i)$. The typically large $D_{\text{LLM}}$ dataset could be used to compute a precise but possibly biased estimate $\hat \theta$ of a target parameter $\theta^*$ of the human generating process (e.g., a population mean).  Alternatively, only the human gold standard labels in $D_{\text{shared}}$ could be used, but this estimator fails to exploit information that the LLM predictions may provide about the target parameter of $\humandgp$. 
Instead, a statistical calibration approach combines information in $D_\text{LLM}$ with that in $D_{\text{shared}}$. Because $D_\text{shared}$ contains both observed and predicted outcomes for the same individuals, it is possible to assess the LLM's errors in predicting human responses and estimate a rectifier term, $\hat \Delta$. The rectifier is then used to calibrate the base estimator, which is either the LLM-only or the human-only estimator depending on the approach. 
Statistical calibration approaches combine these to construct valid (i.e., consistent and asymptotically unbiased) estimators of the target parameter $\theta^*$:
\[
\hat{\theta}_{\text{corrected}} = \hat{\theta}_{\text{base}} + \hat{\Delta}
\]
\noindent Importantly, the resulting estimator is valid without having to assume accuracy or unbiasedness of the LLM predictor. Approaches vary in how they construct the base estimator, what they assume about the sampling of instances for human labeling, and how they use $D_{\text{shared}}$ to adjust the base estimator. Figure \ref{fig:calibration} provides a schematic representation of three types of statistical calibration.

Below, for concreteness, we use an example where the target parameter of the human generating process $\humandgp$ is a population mean $\theta^*=\expect{Y}$ marginalized over inputs. We use $\expect{Y|X=x}$ for the conditional mean at a given input $x$. However, both classes of approaches generalize to more complex parameters like regression coefficients.

\subsection{LLM augmented estimators}   
\label{sec:surrogate}
One class of approaches defines augmented estimators that combine the gold-standard human labeled data with the larger ``surrogate'' LLM labeled set through an additive correction term~\citep{broska2025mixed,egami2024using,gronsbell2024another,miao2025assumption,gan2024prediction}.

A first approach seeks to define an estimator that is consistent for the human target estimand (i.e., converges to the true value as sample size grows), but reduces variance based on inclusion of LLM responses in $D_{\text{LLM}}$ in addition to the smaller set of human responses in $D_{\text{shared}}$~\citep{broska2025mixed}. 
This can be achieved by using a prediction powered inference (PPI) framework. PPI is a general statistical framework for combining gold standard observations with predictions from a surrogate model for parameter estimation~\citep{angelopoulos2023prediction,angelopoulos2024ppiplusplus}.  
The underlying intuition is that as long as the surrogate's predictions offer some information about the outcome that is being predicted, incorporating them can allow for more precise estimates than using gold standard measurements alone. 
In the case of AI surrogates, this translates as expecting more precise inferences--i.e., tighter confidence intervals on estimates of means, treatment effects, regression coefficients, or other parameters--whenever an LLM provides some reasonable prediction of human behavior, even if biased or noisy.

PPI assumes that the features $\tilde X_i$ 
in $D_{\text{LLM}}$ refer to different hypothetical respondents, but are drawn from the same distribution as the features $X_i$ %$X_j, j \in 1,...,n$ 
in $D_{\text{shared}}$\footnote{Alternatively, there should be valid importance weighting using known population characteristics.}. It is also assumed that the LLM $\hat{f}$ is independent of $D_{\text{shared}}$ and $D_{\text{LLM}}$, i.e., these samples were not used to train $\hat{f}$.\footnote{Note this is weaker than the No Training Leakage condition discussed above, which rules out overlap with the broader research context.} 
PPI achieves the goal of more precise combined estimates by adjusting an estimator that only uses gold standard human observations, where the form of the adjustment is designed to prevent bias from being introduced. Precision is improved by optimizing the magnitude of the adjustment, and the resulting estimate at least as precise as it would be with human measurements alone when the adjustment is appropriately tuned.

The PPI estimator for the population mean is:

\[
\hat{\theta}_{\text{PPI}}^{\lambda} = \frac{1}{n}\sum_{i=1}^{n}Y_{i} -\lambda\left(\frac{1}{n}\sum_{i=1}^{n}\hat{f}(X_{i}) - \frac{1}{N}\sum_{i=1}^{N}\hat{f}(\tilde{X}_{i})\right) 
\]

\noindent The first term defines the human estimator, $\hat{\theta}^H$. The parameter $\lambda$ is tuned to minimize the variance of the estimator (see the power-tuning procedure in \citep{angelopoulos2024ppiplusplus}), such that $\hat{\theta}_{\text{PPI}}^{\lambda}$ is at least as precise as the human subjects estimator. 
In general, provided $n$ and $N$ are sufficiently large and standard regularity conditions hold, the PPI estimator of a parameter is consistent and asymptotically normal, yielding asymptotically valid confidence intervals. Further improvements to the efficiency of the PPI estimator are explored in recent work~\citep{angelopoulos2024ppiplusplus,gronsbell2024another}.

Exemplifying the PPI approach, \citet{broska2025mixed} define a mixed subjects design approach that applies PPI to \textit{design analysis}. 
One use of their framework is to determine the effective sample size for a mixed design study: the equivalent number of human subjects the researcher would need in order to get an estimator as precise as what they get from the mixed responses. 
Another is to determine the appropriate mix of human and lower-cost LLM subjects, given some research budget, to achieve a chosen power level and effective sample size.

A related approach is termed \textit{design-based supervised learning} (DSL) by \citet{egami2024using}. DSL similarly adjusts a base estimator using information from $D_{\text{shared}}$, but begins from the LLM-only estimate, which is adjusted by subtracting an inverse probability weighted correction using $D_{\text{shared}}$. This is possible under a design-based labeling scheme in which the labeling (inclusion) probabilities are known, possibly as a function of observed covariates. In the simplest case of mean estimation where human labels are collected for a simple random sample of the inputs, the estimator is:

 \[
\hat{\theta}_{\text{DSL}}
= \frac{1}{N}\sum_{i=1}^{N}\hat{f}(\tilde{X}_{i})
- \left(\frac{1}{n}\sum_{i:\,R_i=1}\hat{f}(X_{i})
-\frac{1}{n}\sum_{i:\,R_i=1}Y_{i}\right)
\]

where $R_i$ is an indicator of whether input $i$ is human labeled (i.e., in $D_{\text{shared}}$).
In the case of mean estimation, this correction is equivalent to using the PPI correction when $\lambda=1$.

In practice, social scientists often face a tension between the broader population they wish to generalize to and the more narrowly defined population they obtain human convenience samples from--e.g., undergraduates in a psychology subject pool or online participants recruited through crowdsourcing. While design-based supervised learning  assumes that the probability of labeling a sample is known by the researcher in advance, \citet{guerdan2025doubly} provide a doubly robust estimation framework in which the instances $X$ and $\tilde{X}$ in $D_{\text{shared}}$ and $D_{\text{LLM}}$ respectively are not necessarily drawn i.i.d. from the same target distribution, in addition to the human labels in $D_{\text{shared}}$ not necessarily being missing at random. Here, combining the LLM-augmented estimator with a reweighting estimator yields a doubly robust calibrated estimator that remains valid if either the reweighting model or the outcome model is correctly specified.

\subsection{Learning a plug-in bias correction}
While the previous estimators make use of both the measurements in $D_{\text{shared}}$ and $D_{\text{LLM}}$, a second class of approach uses $D_{\text{shared}}$ to learn the relationship between LLM and human responses, under similar assumptions as above about the distribution from which features $\tilde{X}_i$ and $X_i$ are drawn and the training of the LLM. The resulting model is then used either to adjust the LLM predictions in $D_{\text{LLM}}$ so that they behave like de-biased proxies for human outcomes prior to plugging them into the target estimating equation~\citep{ludwig2025large}, or to adjust the target inference directly~\citep{wang2020methods}.  

\citet{ludwig2025large} describe plug-in bias correction, where $D_{\text{shared}}$ is used to estimate a model $\hat{b}$ that predicts a conditional bias function:
\[
b(x):=\expect{\hat{f}(X)-Y|X=x}
\]
capturing the expected error of the LLM prediction.
For example, the researcher might fit a linear model or a nonparametric learner, using standard methods to avoid overfitting.

For each input $\tilde{X}_j, j=1 \dots N$ in $D_{\text{LLM}}$, they can then construct bias-corrected pseudo-labels, i.e., $\hat{Y}_{j}=\hat{f}(\tilde{X}_j)-\hat{b}(\tilde{X}_j)$.

Assuming $D_{\text{shared}}$ is used only to estimate the bias model, the plug-in estimator for the mean is
\[
\widehat\theta_{\text{plug-in-corrected}}
=
\frac{1}{N}\Bigg(\sum_{j=1}^{N} \big\{\hat{f}(\tilde{X}_j)-\hat{b}(\tilde{X}_j)\big\}
\Bigg)
\]

\noindent Alternatively, one can use a cross-fitting, where the bias is learned on held-out folds of $D_{\text{shared}}$ and then used to de-bias predictions for all instances in $D_{\text{LLM}}$. If the bias function $\hat{b}$ is accurately learned, then as sample size approaches infinity, $\widehat\theta_{\text{debias}}$ will converge to the true value (i.e., the estimator is consistent) and is asymptotically normal under standard regularity conditions. Accurate learning of the bias implies it is not overfit; the fitting process should produce similar results if slightly different samples from the population of instances appeared in $D_{\text{shared}}$. 
Similar to the PPI estimator above, the precision of the corrected estimator $\widehat\theta_{\text{debias}}$ will depend on the size of $D_{\text{LLM}}$ relative to that of $D_{\text{shared}}$, as well as the variability of the target outcome and of the estimated bias. 

Rather than forming pseudo-labels, \citet{wang2020methods} fit a relationship model using $D_{\text{shared}}$

\[
Y \;=\; K\!\big(\hat{f}(X)\big)\;+\;\varepsilon
\]

\noindent For example, this could be a linear model $Y=\alpha+\beta\,\hat{f}(X)+\varepsilon$. This model is used to adjust the downstream inference done on $D_{\text{LLM}}$, correcting coefficients and standard errors. For example, when the target is a population mean and $K$ is a linear model, the estimator becomes $\widehat\theta_{\text{corrected}}
= \hat\alpha+\hat\beta\big(\frac{1}{N}\sum_{j=1}^{N} \hat{f}(\tilde{X}_j)\big)$ .

\subsection{Limitations of statistical calibration}

\subsubsection{Barriers to increasing precision in the behavioral sciences}

To date, empirical applications of statistical calibration suggest that augmenting human data with LLM simulations yields only modest gains in precision. For example, \citet{broska2025mixed} augment a sample of 10,000 human decisions in moral dilemmas with 100,000 LLM-predicted decisions. Although PPI enabled unbiased causal effect estimates for factors influencing these decisions on average, augmenting human sample LLM-predicted decisions increased the effective sample size by only about 13\%, from 10,000 to at most 11,275. Similarly, \citet{krsteski2025valid} apply PPI to two large-scale social surveys using 100 human responses as ground truth, and observe improvements in effective sample size up to 14\%. In what follows, we outline three explanations for these modest gains in precision and draw out implications for future research.

First, the benefits of statistical calibration are constrained by the accuracy with which LLMs predict human behavior. Biases, misrepresentations, and other errors that reduce prediction accuracy limit potential increases in the precision of parameter estimates. Effective sample size and similar metrics of data quality indicate that, at present, LLM predictions of human behavior contribute relatively little information for parameter estimation. Future work could explore ex ante repair strategies, potentially in combination with more capable models, as routes to more accurate predictions, used in conjunction with statistical calibration.

%attitudes
%- measurement error
%- issue salience 
%- changing attitudes over time
%- inconsistent answer

However, even with improved models, limits to predictability arise from the nature of human data. For example, knowing about a person's attitude on one sociopolitical issue does not imply we can successfully predict another attitude because only a minority may hold internally coherent views on sociopolitical issues and be able to articulate them when asked \citep{converse2006beliefsystems, freeder2019importance}. Numerous examples from surveys and behavioral experiments show fluctuations in responses given by the same individual to the same question\citep[e.g.,][]{dean1977measuring,zaller1992simple,wikman2006reliability}. Prediction errors may cluster around specific cases or individuals and resist accurate recovery \cite{salganik2020fragile}. Under such conditions, not only LLMs but also prediction algorithms more broadly are likely to yield limited gains in predicting human behavior.\footnote{By contrast, prediction algorithms may yield greater benefits when applied to text, images, and other data with a higher signal-to-noise ratio \citep{gligoric2025can,maranca2025correcting}. 
These annotation tasks typically allow some inter-rater disagreement but assume it is limited.} More research is needed to characterize the predictability of human responses in experiments and to assess how closely LLMs and other algorithms can approach this ceiling.

Finally, the limited increases in precision can be attributed in part to the mechanics of the calibration methods themselves. Existing methods do not always make the most efficient use of the available data for estimating parameters. For example, the original formulation of PPI \citep{angelopoulos2023prediction} has been superseded by PPI++, which offers greater increases in precision \citep{angelopoulos2024ppiplusplus}. This is an area of active development (see \citep{hoffman2024we} for a recent review), suggesting considerable potential for further improving statistical calibration methods. These methodological advances, together with improvements in predictive accuracy and a clearer understanding of the predictability of behavior through LLMs, will determine how far statistical calibration can take researchers towards valid inferences of causal effects and other parameters.

%On the bright side, modest observed gains in precision reduce the risk that silicon samples will make us much more confident in estimates that are biased for traditional, human reasons, such as unrepresentative human samples~\citep{henrich2010weirdest,yarkoni2022generalizability}.
%While the true data-generating process for human behavior is highly complex, high levels of noise in measurements may lead to scenarios in which it is hard to greatly improve prediction over simple models. A related factor concerns how behavioral studies are designed by researchers. For example, \citet{morucci2024model} argue that many datasets related to political behavior have a ``low intrinsic dimension'' based on how they are collected by researchers who think in fundamentally linear ways when hypothesizing about which variables relate to the target outcomes. \jessica{wrote this quickly, can be improved}

%While the above approaches will produce more informative estimates than substituting LLM responses for human ones based on presumed interchangeability, 
\subsubsection{Potential to overfit bias estimates}
When using calibration approaches researchers should keep in mind that no method can guarantee unbiased estimates for a single finite sample. Our own experience discussing these methods with researchers suggests to us that this is not always well recognized. Like any approach to measurement error modeling and correction, calibration approaches hinge on having enough data to reliably estimate the bias contributed by using LLM responses. 
%Even when using PPI to estiamt, where the expected value of the estimator will equal the true target value (i.e., it is finite sample unbiased), 
%It is important that researchers recognize that guarantees, such as unbiased assume repeated sampling. Guarantees  As a statement about what happens in expectation, unbiasedness cannot be guaranteed for any fixed finite sample. 
This leads to an important question: 
Once the researcher has collected enough data for stable bias estimates, would they be just as happy using human responses alone? 
If the resulting gains in precision are small or do not qualitatively change the conclusions---for example, because the human sample size needed to estimate bias already delivers adequate power and statistical significance---the added complexity of integrating LLM samples may not seem worth the trouble. On the other hand, for high stakes topics with strong incentives towards precision under budgetary constraints (e.g., health interventions), a mixed subjects framework may provide a more natural step toward further optimization.

\begin{figure}
    \centering
    \includegraphics[width=0.7\linewidth]{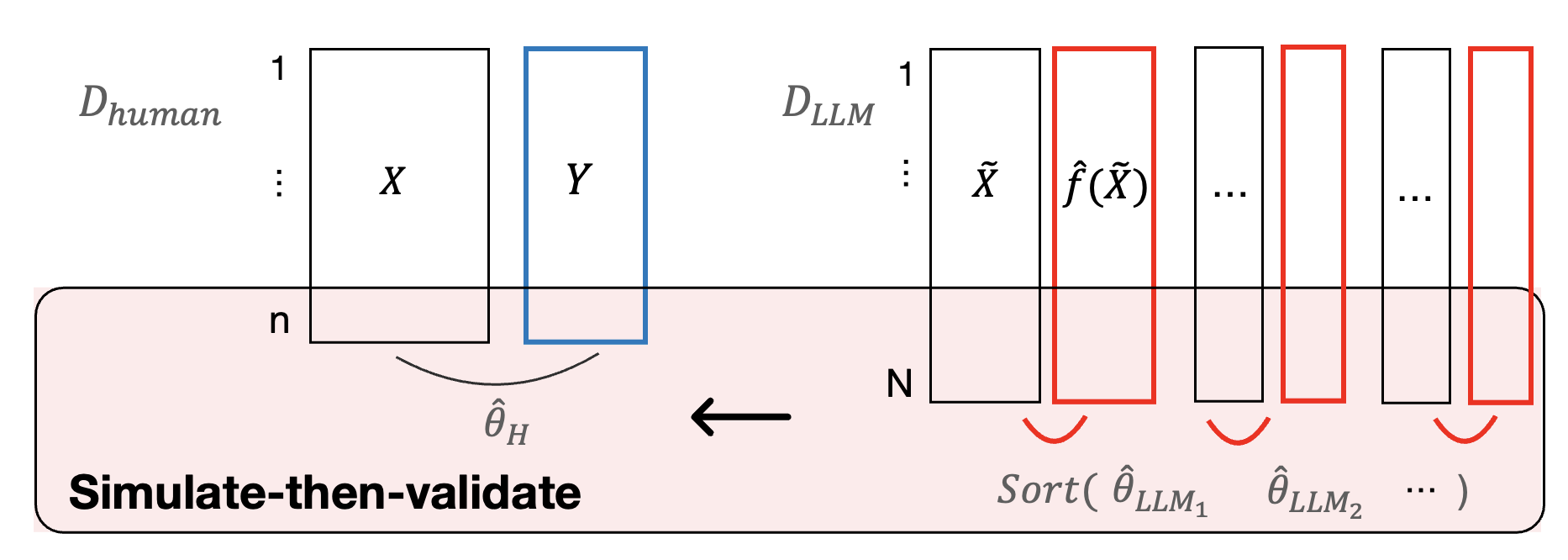}
    \caption{In a simulate-then-validate approach, the researcher uses LLMs exclusively for exploratory piloting to find hypotheses with support, then conducts a human study on those that appear most promising.}
    \label{fig:simulate-then-validate}
\end{figure}

\section{Language models and the role of simulation in behavioral science}
\label{sec:discussion}

In contrast to using statistical methods that account for added bias, some have proposed that a lower risk alternative is to use LLMs in preliminary phases of research, such as in exploratory piloting to discover effects. 
This raises questions about how to best leverage the signal LLMs provide to make preliminary research more efficient, and how to evaluate their impacts on fundamentally open-ended pursuits like generating hypotheses or identifying outcomes.
%We discuss applications of LLMs to discovery, design analysis, and hypothesizing mechanisms to explain human behavior. 

\subsection{Simulate-then-validate: Using AI surrogates to rank candidate designs}
One way to avoid the stringent conditions required for plug-in use of LLMs for inference is a \textit{simulate-then-validate} approach \citep{kozlowski2025simulating} (Figure \ref{fig:simulate-then-validate}). 
Here, AI surrogates are used as a discovery tool: models are used to simulate running many possible study designs to identify hypotheses with support, and only those that show promising results with AI surrogates are subsequently evaluated with human samples. For example, authors have argued that ``at a minimum, studies that use simulated
participants could be used to generate hypotheses''~\citep{grossmann2023ai}. Related proposals promote LLMs as a means of increasing design efficiency by pretesting and refining attention checks, reverse-coded items, branching logic, and outcome measures before exposing human subjects~\citep{grossmann2023ai,sarstedt2024using}. %In other cases, models are used to critique and extend study materials, such as by identifying ambiguity in prompts, suggesting simplified wording, or producing images~\citep{olivos2025chatgptest,valenzuela2025using}. 

Researchers have used LLMs to estimate target effects consistent with relevant domain knowledge prior to running human studies. For example, \citet{manning2024automated} provide an LLM with a structural causal model that defines which variables should be manipulated, and use it to simulate many experimental trials consistent with this generating model. The effects that are estimated (under the assumed causal model structure) can be used identify where a potentially interesting association between an outcome measure and experimental scenario may exist with humans.

The simulate-then-validate proposal speaks directly to several challenges in exploratory research. Consider the distinction between \emph{theory-testing} research, in which a theory implies that a specific hypothesis should hold, and \emph{discovery-oriented} research, in which a theory merely defines a broad class of possible effects \citep{oberauer2019addressing}. In the latter case, the truth of any particular hypothesis is a priori unlikely; in other words, the base rate $P(\hyp \mid \theory)$ that a randomly chosen hypothesis $\hyp$ implied by theory $\theory$ is true is low, even if the theory is broadly reasonable. Yet if $P(\hyp \mid \neg \theory)$ is much smaller than $P(\hyp \mid \theory)$, then finding one such hypothesis with empirical support still raises the posterior probability of the theory. The researcher faces a search problem: the space of possible effects is typically large, and only a small subset of possible hypotheses are true. If this were not the case, positive findings would not be considered so scientifically interesting. 

%LLMs appear to offer leverage on this point, as they allow the researcher to search a larger portion of the space of possible effects than is possible with humans. 
Imagine a researcher who speculates that mindfulness interventions can increase student's proclivity to study by improving perceived self-control. Of the many ways one could design a study to intervene on mindfulness and operationalize proclivity to study, we expect only a small fraction of those designs to in fact yield empirical regularities that fit the theory.
Using AI surrogates to screen designs and prioritize those with larger or more consistent apparent effects could, in principle, improve the efficiency of this search and even turn up larger effects than would be found under more limited human piloting. LLMs may also be useful to brainstorm possible manipulations or structural models based on the latent information about human behavior that they encode.%, (see, e.g., the use of LLMs to generate outcomes of interest and their potential causes in \citet{manning2024automated}). 

Using AI surrogates to discover effects also targets a second challenge faced in exploratory research: the high risk of false positives that is typical in pilot studies.  Human pilots often rely on small samples. When statistical significance is used as a filter to identify effects worthy of follow-up research, lower statistical power increases the chances that a result that passes the significance threshold is a false positive~\citep{button2013power}. Hence, a discovery-oriented activity like piloting often produces noisy results for purposes like estimating average treatment effects~\citep{gelman2024causal} or deciding which treatment variation to carry forward. 
%Naturally, smaller samples also lead to a higher risk of false negatives.
LLMs make it possible to cheaply generate large numbers of measurements on the same design, albeit at the risk of introducing model-based bias.

At the same time, the low base rate of often small target effects that motivates discovery-oriented search also limits the transformative potential of the simulate-then-validate approach. 
%If all possible interventions produced effects, finding evidence of an association is less likely to be deemed a surprising new result.\footnote{\jessica{Comment on critiques of fields like psychology's preoccupation with counterintuitive effects}} 
%Imagine the researcher runs pilot versions of several of the many possible studies, and finds that one produces a positive effect in line with the theory. 
 %It is researchers' implicit expectations about low base rates for new behavioral results explain why a new result is considered ``newsworthy.'' 
%When LLMs are used in place of human participants for all or any portion of what would be human pilot results, it becomes imperative that they accurately predict counterintuitive human behaviors.  \jessica{summarize any existing evidence on how well they have predicted surprising results versus things that are well predicted by the literature by now}
Many social science studies target relatively small effects, in the order of around $r=0.1$ to $r=0.35$~\citep{funder2019evaluating,schafer2019meaningfulness}. Searching for hypotheses with empirical support therefore means searching a space with low base probability of a \textit{small} effect among humans. %When researchers use significance testing as a decision rule and can exploit design degrees of freedom~\citep{gelman2013garden}, reported test statistics tend to bunch close to the decision boundary, as demonstrated by analyses across economics, psychology, political science, and other disciplines~\citep{brodeur2016star,gerber2008statistical,hartgerink2016distributions,stommes2023reliability}. 
Even small amounts of bias in LLM predictions in the piloting phase could lead to high rates of false positives in piloting. As we describe above, recent work finds that LLMs tend to overestimate effect sizes and to produce high rates of statistically significant effects even for designs that yielded non-significant findings in human samples~\citep{cui2025large}. Even with very large numbers of simulated trials, a model's ability to predict the sign of human effects likely depends on how far those effects are from zero. Corroborating this, \citet{manning2024automated} find that across four example experiments and 12 total causal paths exploring social interaction scenarios (e.g., bargaining over a mug, ascending price auction), the LLM-attained estimates correctly predicted the sign 10 times, with the failures occurring for two of the smaller effects studied. Proposals to use LLMs to identify boundary conditions delineating when the predictions of a theory are likely to fail~\citep{tranchero2024theorizing} may similarly struggle in practice as a result of systematic bias.

LLM's tendency to overestimate true behavioral effects leads to proposals that they be used primarily to gain \emph{relative} information, e.g., to identify the top-performing intervention rather than to rely on absolute effect size estimates~\citep{sarstedt2024using}, analogous to how ``mega-studies'' with many conditions are used \citep[e.g.,][]{voelkel2024megastudy,milkman2022680, patel2023randomized}. The challenge that arises is that the information provided by LLM rankings will be coarser than the ranking implies. Determining how much coarser for a given domain requires trial and error in comparison with human data. 
If labs got in the practice of transparently reporting how they used LLMs in discovery, along with final human results, the question of ``how much coarser'' could be answered empirically.

Of course, the simulate-then-validate approach need not be restricted to planning novel studies. Use of LLM simulation studies could also be routinized to prioritize which previously discovered effects are most worth conducting a replication study. 
% The chunk below was migrated from the intro
Social phenomena can change over time, false-positive results can occur, and methodological limitations can undermine confidence in original results. While the scientific literature is vast, resource limitations have led to replication studies being reserved for a very small proportion of published results. LLMs make it easy for behavioral scientists to recreate the original experimental conditions and outcome measures in a simulation study, check whether the reported effect reproduces under reasonable assumptions, and then prioritize replications with human samples for findings that appear fragile or inconsistent.

\subsection{Causal discovery of features and outcomes}
Another promise of LLMs for behavioral science is not just cheaper measurement, but cheaper search over what to measure, including outcomes and inputs. For example, recent work suggests LLMs can accelerate \emph{outcome discovery} in text-based behavioral studies by first testing whether treatment assignment is predictable from text on held-out data, then using an LLM to propose a small set of human-interpretable themes and scoring rubrics~\citep{modarressi2025causal}. A completeness measure benchmarks theme-based prediction against an unconstrained predictor to determine how much of the detected difference is captured by the focal themes. 
A complementary line of work targets \emph{feature-level causal effects}, estimating the causal effect of treatments instantiated in unstructured text on human behavior by accessing and manipulating the model's internal activations to disentangle the treatment features of interest from other possible confounding features~\citep{imai2024causal}. These approaches demonstrate how principled use of LLMs can transform exploratory hypothesis generation as well as estimation of effects.

\subsection{Design analysis and stress-testing}
%While the prevailing proposal for using LLMs in exploratory research treats them as stand-ins for human responses, researchers may also benefit from simulations to stimulate their imagination during design and interpretation. In this view, rather than looking to LLMs to provide forecasts of what will happen with human subjects, they are used to think through possibilities. For example, 
Robust study design and evaluation depends on being able to produce reasonable estimates of target effects. 
For example, effect size estimates are used to determine how many subjects a design should run to achieve reasonable statistical power to detect a target effect. 
%In consuming results, it is often useful to contextualize interpretations of effect size estimates for decision-making against expectations based on domain knowledge or prior work, to help guard against being misled by spurious findings. 
%Many researchers may not be comfortable coding their own generative simulations of data or otherwise interrogating estimates of effect size.  
%Consequently, estimates used to power studies can be unrealistic, resulting in underpowered designs, and published effects can often appear too good to be true. 
%Language models may be useful as dialogue assistants for the purposes of design and analysis interrogation. An LLM could play a role similar to that of a consulting statistician, prompting the researcher to reflect on domain knowledge and challenging their initial estimates or interpretations of effect sizes through simulation. 
By enabling simulation of possible data scenarios through natural language descriptions, language models lower the barrier to simulation-based design analysis. For example, an LLM could take information about the expected scale and variance of measurements and use it to generate simulated datasets that align with versus that challenge the researcher's expectations, or that help them consider what their effect size estimates might imply about possible distributions of individual-level effects~\citep{gelman2024causal}. LLMs' ability to generate predictive distributions based on prior data points and textual descriptions provided via prompting~\citep{requeima2024llm} makes them a good first pass tool for producing distributional expectations, which the researcher could refine through natural language dialogue or providing additional data points capturing their expectations. 
In planning analysis, simulating data that violates assumptions (e.g., about response scales) or varies in structure (e.g., from little to high inter-variable correlation) provides a preliminary check of the robustness of an analysis plan to deviant data; LLMs could help here as well. Starting points for such approaches might be found in existing LLM surrogate tools \citep[e.g.,][]{park2024generative, toubia2025database} or even adversarial ``red teaming'' systems such as the synthetic survey response agent developed by \citet{westwood2025potential}.

\subsection{Hypothesizing mechanism}
%Many applications of language models in behavioral research treat them as simulated subjects while making minimal changes to the process that would be used to prompt human participants. This trend fails to make use of the unique access LLMs can provide to internal representations that impact their behavior. While notoriously referred to as a black box, an LLM is less of a black box than a human for some purposes. 
Consider the common practice of asking study participants why they responded in a certain way to a stimulus, or what strategy they used to arrive at a belief or decision. While we can elicit such explanations, we cannot ``open up the hood'' on a human participant to verify that the strategies they report actually affect their behavior. We have long known that human rationales often deviate from true causal descriptions of the participant's process~\citep{nisbett1977telling}. The natural language interface of the LLM, on the other hand, can be sidestepped by accessing its internal representations directly. %Linear probing involves training a ``decoder'' model to predict a target outcome, such as a decision or belief report, from the activations in a layer of the model. Probes provide a way of estimating how much relevant information a model has about a target outcome, though like any model they require care not to overfit~\citep{}. 

For example, sparse autoencoders (SAEs) are neural networks that can be used to extract meaningful features of a text for understanding a model's behavior. The challenge SAEs overcome is that the same neuron in a neural net can control multiple seemingly unrelated concepts. SAEs disentangle representations by 
mapping model activations to an overcomplete representation to which a sparsity constraint is applied, making it possible to identify a small set of learned features (or \textit{concepts}) that are activated for any given text input. %Thus SAEs can be used to generate hypotheses about how an agent is responding to or processing a given text. 
Researchers can fine-tune language models to simulate human behavior, then use SAEs and related interpretability techniques to help them generate theories of what human participants may be reacting to in stimuli. Such approaches can stimulate ideas about follow-up human studies aimed at disentangling mechanism. 
While still nascent in human behavioral science applications of LLMs, demonstrations of probing and SAEs in the machine learning literature show how LLMs can be used to corroborate human-oriented theories, e.g., of emotion~\citep{tak2025mechanistic,lee2025can}.

Models can also be fine-tuned on human behavioral data to generate reasoning traces that explain human strategies, avoiding the need to probe the model internals. For example, \citet{zhu2025using} employ reinforcement learning with outcome-based rewards to guide LLMs toward generating explicit Chain-of-Thought reasoning traces for explaining human risky choices. Relative to a base model, this approach amplifies the model's proclivity to produce explanations that explain human choice behavior, such as references to cognitive biases or psychological factors, which can serve as input to %. Clustering and interpreting explanations to identify common strategies provides fodder for designing 
follow-up studies or theories focused on mechanism.   
\section{Conclusions}
Questions about validating LLM simulations arise in the midst of a tension between what LLMs promise---flexible predictors of plausible text---and the heavy emphasis on confirmatory goals in behavioral studies. 
Until recently, getting good predictions for how people would respond to novel survey items or behavioral experiments required gathering some amount of human data and  analyzing it directly as pilot data or using it to train a supervised machine learning model. Or, it required substantial theorizing or empirical observation to drive the development of cognitive models. 
LLMs promise a general purpose ``crystal ball'' that can produce predictions to any questions without the need for domain specific data. 
The danger is that researchers overgeneralize, treating these results as a stand-in for human responses. 
Proposals to codify heuristic validation~\citep{aher2022using,argyle2023out,tranchero2024theorizing} suggest that the potential consequences of subjecting inferences to unknown bias are not well recognized and/or deemed important.

We argue that while researchers \textit{should} be examining how LLMs do and do not mimic human behavior to better understand opportunities LLM use affords, we must draw a clear line between heuristic use of AI surrogates and methods with calibration guarantees. 
Behavioral researchers routinely accept that statistical conclusions are only credible under explicit assumptions, ranging from randomized (or otherwise ignorable) treatment assignment for causal inference to mean-zero errors uncorrelated with regressors in ordinary least squares regression. No leakage between test and training data and the preservation of critical moment conditions for parameter identification are the necessary conditions for valid use LLM simulations as a plug-in inferential method. Heuristic approaches conflict with standard scientific norms dictating that the validity of behavioral results should not rest on face plausibility alone.

Using AI surrogates in exploratory research to pre-test designs carries less risk, though efficiency gains are not guaranteed. Further corroboration of results is needed on a domain-specific basis to judge how reliable AI surrogates are for ranking experimental scenarios. Thinking outside the ``silicon sample'' box to consider using models for causal discovery of outcomes and features and to explore possibilities in design analysis may be just as fruitful. Rather than seeing LLMs only as tools to generate plausible human responses, other features they offer behavioral scientists can be leveraged to improve design and theory, such as their flexible natural language interface for data simulation without programming expertise, and ability they provide to access and manipulate their internal representations. Exploratory and design-focused applications of LLMs may also introduce their own risks, however, insofar as they entail the use of LLMs for decision-support, thereby creating further opportunities for bias or inefficiency \citep{vaccaro2024when}.

Our argument is premised on the idea that so long as we rigorously account for differences between LLM and human responses in inference, we can learn from LLM responses regardless of their black box nature. However, some have argued that there may be qualitative differences in behavioral research if AI surrogates are heavily adopted. 
One concern is selection effects as researchers gravitate to questions that are conducive to using LLMs. 
The less highly specified the target of the study, the further the LLM approximation may be from human reality, stimulating more focus on what can be learned through highly structured forms of behavioral elicitation.  
Surveys, where the answers to ``who, what, when, where'' are tightly specifiable, can be contrasted with less structured studies like field experiments, which involve uncontrolled, often unobserved, contingencies. Similarly, we would expect LLMs to be better at simulating text-based reactions to messages, e.g., that are meant to persuade in political or marketing scenarios, than at anticipating the effects of paralinguistic signals in interaction (e.g., tone, gesture)~\citep{kozlowski2025simulating}. 
Researchers may gravitate toward domains with lower relative variance in responses (e.g., cognitive memory tasks) compared to higher variance domains (e.g., social psychology). 
This becomes a concern if it weakens the link between questions that are socially or policy relevant versus those to which researchers attend. 
By analogy, survey experiments have enabled effective testing of theoretical mechanisms, but the availability of cost-effective data collection methods like crowdsourcing platforms may have narrowed the space of research questions considered to those addressable with survey experiments~\citep{anderson2019mturkification}.
\citet{messeri2024artificial} summarize this as a risk that LLMs will provide an ``illusion of exploratory breadth,'' a perceived sense of comprehensiveness where scientists falsely believe they are exploring the full space of testable hypotheses, whereas they are actually exploring a narrower space. Ultimately, as is the case for any method, researchers must form their own views on the potential implications of widespread use of that method for progress in their community of interest.

%As a thought exercise, consider a researcher interested in interventions for improving people's self-perceptions of control over their physical activity, which prior research showed to correlate with physical activity levels. They imagine that mindfulness training will boost the subject’s perceived ability to control their physical behavior ($\theory$), resulting in a specific hypothesis $\hyp$ predicting that subjects who do a two minute guided mind-centering exercise will report higher proclivity to exercise in the next hour than subjects who watch a relaxing video that doesn't contain mindfulness training. Given that an LLM cannot control its physical realization, does it ever make sense to compare or treat as commensurate LLM results $\resultfun_{LLM}(x)$ and human results $\resultfun_{H}(x)$?

%\begin{figure*}[h]    \includegraphics[width=\linewidth,height=0.2\textheight]{example-image-a}
%    \caption{Figure 1}
%    \label{Fig::Overview}
%\end{figure*}

%\input{sections/03_preliminaries}

\clearpage
\nottoggle{blind}{%
%\subsection*{Acknowledgements}
%This work was supported by funding from the National Science Foundation IIS 2040880, IIS 2313105, and the NIH Bridge2AI Center Grant U54HG012510.%
}

\iftoggle{icml}{%
\bibliographystyle{ext/icml2025.bst}
}{
}
{\small\bibliography{llm_behavioral}}

%\iftoggle{neurips}{\input{neurips_2025_checklist}}{}

%%% Appendices

\section{Appendix}
\label{sec:appendix}

\subsection{Research context and substitution requirements \citep{ludwig2025large}}

A \textit{research context} specifies how text strings are sampled for the study and for model training. Let $\Sigma^*$ be the space of strings that may be presented as prompts (scenarios). Given analysis inputs $X=(S,A, W)\in\mathcal X$, assume $S\in\Sigma^*$ is the prompt string and $Z=(A,W)\in\mathbb R^k$ are structured covariates (treatment and subject-level observables) used in analysis. Assume the same $\Sigma^*$ governs strings that may appear in the model’s training data $t$. 

Let $C=(C_\sigma)_{\sigma\in\Sigma^*}$ and $T=(T_\sigma)_{\sigma\in\Sigma^*}$ be \emph{string-level} inclusion indicators: $C_\sigma\in\{0,1\}$ indicates whether string $\sigma$ is included among the prompts used in the study; $T_\sigma\in\{0,1\}$ indicates whether $\sigma$ appeared in the model’s training data $t$. Equivalently, we can write the training indicators as the vector $\train = (T_\sigma)_{\sigma\in\Sigma^*}$. A research context $Q(\cdot)\in\mathcal Q$ is a joint distribution over $(C,T)$. We use $Q$ only for string-level statements; all inference statements below take expectations over a distribution $P_X$ on $X=(S,A,W)$.

\citet{ludwig2025large} make two assumptions on the sampling of strings in a research context. All research contexts satisfy:
\begin{enumerate}
  \item For any realizations $c,t$, $Q(C=c,T=t)=\prod_{\sigma\in\Sigma^*} Q(C_\sigma=c_\sigma,\,T_\sigma=t_\sigma)$.
  \item The marginal number of study prompts is unaffected by conditioning on $T$: $\mathbb{E}_{Q}\!\big[\sum_{\sigma\in\Sigma^*} C_\sigma\big]=\mathbb{E}_{Q}\!\big[\sum_{\sigma\in\Sigma^*} C_\sigma \,\big|\, T=t\big]$.
\end{enumerate}

Additionally, the LLM $\llm{.}{\train}$ is assumed to be deterministic; i.e., it is a blackbox text generator that takes in a text string $\str$ from an alphabet $\Sigma^*$ and greedily outputs its most likely response.

\subsubsection{Requirements for LLM substitution}

Two high-level properties are sufficient for treating an LLM as an interchangeable proxy for human measurement within a research context.

\paragraph{Condition 1: No Training Leakage.}
Because leakage pertains to the \emph{prompt string}, the relevant indicators for the research context are at the string level. Let $C_S:=C_{S}$ and $T_S:=T_{S}$ denote the inclusion indicators for the realized prompt $S$. To obtain unbiased estimates of the LLM’s predictive performance on the scenario population, the model must not have been trained on any string that enters the study:
\[
\Pr_{Q}\!\big(T_S=1 \wedge C_S=1\big)=0 \quad \text{for all } S\in\Sigma^*,
\]
where the probability is taken under the research context $Q$.

\paragraph{Condition 2: Preservation of Necessary Assumptions for Parameter Identification.}
Downstream inferences about behavioral parameters are valid under substitution only if the LLM preserves the \emph{analysis moments} at the conditioning level used in the estimator. Let $V(X):=\mathbb{E}[Y\mid X]$ and, when conditioning only on the prompt, $V_S(s):=\mathbb{E}[Y\mid S=s]=\mathbb{E}[V(X)\mid S=s]$. Even when pointwise prediction error is small, e.g.,
\[
\big|\,\hat{V}(x)-V(x)\big| < \varepsilon \quad \text{for all } x\in\mathcal X,
\]
parameter estimates can be biased if LLM errors vary systematically with covariates used in the analysis.

If human labels were observed for the study inputs $\{X_i\}_{i=1}^N$, the researcher would estimate a parameter $\theta\in\Theta$ by solving a sample analog of a moment criterion:
\[
\hat\theta \;=\; \arg\min_{\theta\in\Theta}\; \frac{1}{N}\sum_{i=1}^{N} g\!\big(V(X_i),\,X_i;\,\theta\big),
\]
where $g(\cdot)$ identifies the parameter at the population level.

Under substitution, the researcher plugs in the LLM-predicted measurement:
\[
\hat\theta \;=\; \argmin_{\theta\in\Theta}\; \frac{1}{N}\sum_{i=1}^{N} g\!\big(\hat V(X_i),\,X_i;\,\theta\big).
\]
For the substitution to be valid, the population moments must be preserved:
\[
\mathbb{E}\!\big[g\big(\hat V(X),X;\theta\big)\big]\;=\;\mathbb{E}\!\big[g\big(V(X),X;\theta\big)\big] \quad \text{for all } \theta\in\Theta,
\]
where expectations are taken over the distribution $P_X$ of analysis inputs.\footnote{\citet{ludwig2025large} make several additional assumptions of the moment condition $g(\cdot)$ to derive these results: first, that the moment condition is differentiable in $\theta$ with uniformly bounded derivative and non-singular Jacobian at $\theta^*$ (for identification), and second that the moment condition is sensitive to $V(X)$, by requiring a strictly positive lower bound on the partial derivative of $g(\cdot)$ with respect to the plugged-in measurement $v$ for all $v$ and $\theta$.}%

%\footnote{\citet{ludwig2025large} also assume standard regularity for the moment problem: (i) differentiability of $g$ in $\theta$ with uniformly bounded derivative and nonsingular Jacobian at $\theta^\*$ (identification), and (ii) outcome sensitivity, i.e., there exists $\bar G>0$ such that $\inf_{v,\theta}\big\|\partial g(v,x;\theta)/\partial v\big\|\ge \bar G$ for $P_X$-almost all $x$.}

\subsection{Alternative causal inference treatment \citep{perkowski2025when}} 
%Assume that the true outcome for an individual under a specific treatment can be decomposed into (i) an individual-specific baseline component, (ii) a treatment effect, and (iii) random idiosyncratic error. 
%The LLM’s simulated outcome is assumed to equal the true outcome plus two additional, structured errors: an \emph{individual-specific} prediction bias that is the same under both treatment prompts (e.g., systematically worse predictions for some subpopulations), and a \emph{treatment-specific} (prompt) bias (e.g., one treatment happens to be described in a way that is closer to the model’s training distribution), plus residual variation in the LLM response.
%Under this additive error structure, individual-specific bias cancels when comparing the LLM’s treatment and control predictions for the same individuals. As a result, valid estimation of a constant ATE is possible provided the LLM does not systematically favor one treatment over the other; i.e., the prompt-induced bias is (on average) the same in treatment and control. When they do interact, such that some individuals are more or less sensitive to prompt-related biases, this cancellation can fail, and stronger conditions must hold.
%Here valid inference of constant treatment effects is possible when on average, the interaction between individual-level bias and differences in prompting bias across treatments in the subject population are zero.\footnote{For heterogenous treatment effects, the residual noise term must also be mean zero conditional on the covariate profile.}. 

\citet{perkowski2025when} formalizes requirements for valid causal inference in ``digital twin'' experiments in a Neyman--Rubin potential-outcomes framework, where the treatment is binary $Z=\{0,1\}$ and the estimand is the (constant by default) human average treatment effect (ATE) ($\expect{Y_i(1)-Y_i(0)}$). 

Human potential outcomes follow an additive model,
\[
Y_i(Z) \;=\; \theta_i + Z\tau + \varepsilon_i(Z),
\]
where $\tau$ is the constant treatment effect $Y_i(1) - Y_i(0)$, $\theta_i$ represents stable, individual-specific characteristics and $\varepsilon_i(Z)$ is an idiosyncratic mean-zero noise term. The LLM (digital twin) is modeled as producing a noisy approximation to each human potential outcome,
\[
\widehat Y_i(Z) \;=\; Y_i(Z) + \eta_i + \beta_Z + \xi_i(Z),
\]
where $\eta_i$ is an individual-specific simulation bias that does not vary across treatment arms, $\beta_Z$ is a treatment-specific (prompt-induced) bias, and $\xi_i(Z)$ is a mean-zero noise term capturing residual variation in the LLM's response.

Under this model and assumptions, the individual-specific bias $\eta_i$ cancels when comparing the LLM’s treated and control simulations for the same individuals, such that unbiased ATE estimation is possible so long as as the LLM does not systematically favor one condition over the other:
\[
\beta_1 = \beta_0,
\]
called the Treatment-Invariant Simulation Assumption (TISA).\footnote{For heterogeneous treatment effects, an additional requirement is that the residual noise term must also be mean zero conditional on the covariates.}

\citet{perkowski2025when} also analyzes an interaction model in which individual-specific prediction bias and treatment-specific prompting biases do not separate additively, e.g.,
\[
\widehat Y_i(Z) \;=\; Y_i(Z) + \eta_i \beta_Z + \xi_i(Z).
\]
In this case, TISA is no longer sufficient. Identification additionally requires that the population-average interaction between individual-level bias and treatment-specific prompt bias vanishes, i.e.,
\[
\mathbb{E}\!\left[\eta_i(\beta_1-\beta_0)\right] = 0,
\]
termed the Interaction-Invariant Simulation Assumption (IISA).

\subsection{Example of correlated bias in downstream estimation}

Imagine a study where the target analysis regresses an outcome variable (e.g., donation amount, probability judgment, response time, etc.) on a persona indicator like political affiliation:
\[
Y \;=\; \beta_0 + \beta_1 Z + \eta,
\qquad Z\in\{0,1\} ,\quad \mathbb{E}[\eta\mid Z]=0.
\]
In the LLM-substitution setting we observe the proxy $\hat{Y}=Y+\varepsilon$, where
\(\varepsilon:=\hat Y-Y\) is the LLM measurement error.

Assume the LLM is slightly generous for \(Z=1\) and slightly harsh for \(Z=0\), with
\[
\mathbb{E}[\varepsilon\mid Z=1]=+\tfrac{\delta}{2},\qquad
\mathbb{E}[\varepsilon\mid Z=0]=-\tfrac{\delta}{2}
\]
and a balanced design \(P(Z=1)=P(Z=0)=\tfrac12\), so that \(\mathbb{E}[\varepsilon]=0\) overall.
Imagine the researcher runs the same regression with \(\hat Y\) in place of \(Y\):
\[
\hat Y \;=\; \tilde\beta_0 + \tilde\beta_1 Z + \text{residual}.
\]
By the large-sample limit of the OLS estimator,
\[
\tilde\beta_1
\;=\;
\beta_1 \;+\; \frac{\mathrm{Cov}(Z,\varepsilon)}{\mathrm{Var}(Z)}.
\]
Here \(\mathrm{Var}(Z)=\tfrac14\), while
\[
\mathrm{Cov}(Z,\varepsilon)
= \mathbb{E}[Z\varepsilon]-\mathbb{E}[Z]\mathbb{E}[\varepsilon]
= \tfrac12\cdot \tfrac{\delta}{2} - \tfrac12\cdot 0
= \tfrac{\delta}{4}.
\]
Therefore
\[
\tilde\beta_1
\;=\; \beta_1 \;+\; \frac{\delta/4}{1/4}
\;=\; \beta_1 + \delta.
\]

Hence, even with \(\mathbb{E}[\varepsilon]=0\) overall, small treatment-aligned errors shift the
regression slope.

\subsection{Estimating PPI parameter $\lambda$ for a population mean}

For a population mean, $\lambda$ is estimated as:

\[
\hat{\lambda} = \frac{N}{N+n}
\frac{\widehat{\text{Cov}}(Y_i,f(X_i))}{\widehat{\text{Var}}(f(X_i))}
\]

Here, $\frac{N}{N+n}$ is a shrinkage factor, which gives greater weight to the unlabeled data when $N \gg n$. If $N \le n$, then  $\frac{N}{N+n}$ is at most $\frac{1}{2}$, bringing $\hat{\theta}_{\text{debias}}^{\lambda}$ closer to the human subjects estimator. 
The ratio $\frac{\widehat{Cov}(Y_i,f(X_i))}{\widehat{Var}(f(X_i)}$ is the least squares coefficient from regressing $Y_i$ on $f(X_i)$. As the numerator grows, meaning $f(X_i)$ explains more variance in $Y_i$, $\hat{\lambda}$ is larger and the variance of $\hat{\theta}_{\text{debias}}^{\lambda}$ will be smaller than that of the human subjects estimator $\hat{\theta}^H=\frac{1}{n}\sum_{i=1}^{n}Y_{i}$.
%The form of the estimator ensures unbiased estimates of $\theta^*$ for any prediction algorithm $f$ and any choice of $\lambda$.
Assuming adequate sample sizes $n$ and $N$, the difference in the two LLM-based estimates (the second and third term in the PPI estimator equation in Section~\ref{sec:surrogate}) has expectation 0, so that the distribution of $\hat{\theta}_{\text{debias}}^{\lambda}$ is centered at the parameter targeted by the human subjects estimator, with precision that is no worse (and often better) than the human subjects estimator. This is true for any prediction algorithm $f$ and $\lambda$.

\subsection{Estimating bias for plug-in correction}

Let the instance-level prediction error be $\Delta(X):=f(X)-Y$. 
The goal is to learn a (possibly conditional) \emph{bias function} to estimate this quantity:
\[
b(x)\;:=\;\expect{\Delta(X)\mid X=x}.
\]

When the target parameter is the population mean $\theta^*=\expect{Y}$, we can write $\theta^* = \expect{Y} = \expect{f(X)}-\expect{b(X)}$, by the fact that $Y = f(X)-\Delta(X)$ and the law of iterated expectations. The plug-in estimator below approximates this decomposition by replacing $b$ with its estimate and the expectation with an average over $D_{\text{LLM}}$.

The researcher uses $D_{\text{shared}}$ to learn a predictor $\hat{b}$ of $b$. For example, they might fit a linear model or a nonparametric learner, using standard methods to avoid overfitting.

They can then construct bias-corrected pseudo-labels for instances in $D_{\text{LLM}}$:
\[
\hat{Y}_{j}=f(\tilde{X}_j)-\hat{b}(\tilde{X}_j), j=1 \dots N
\]
Assuming $D_{\text{shared}}$ is used only to estimate the bias model, the plug-in estimator for the mean is
\[
\widehat\theta_{\text{debiased}}
=
\frac{1}{N}\Bigg(\sum_{j=1}^{N} \big\{f(\tilde{X}_j)-\hat{b}(\tilde{X}_j)\big\}
\Bigg)
\]

\noindent Alternatively, one can use a cross-fitted version of the bias-corrected plug-in estimator to reduce overfitting, where for each of $K$ folds, the bias model is learned on the other $K-1$ folds of $D_{\text{shared}}$ and applied to the held-out fold and to debias the predictions in $D_{\text{LLM}}$. The final bias corrected plug-in estimator can then be found by aggregating the $K$ fold-specific estimates.

If the bias function $\hat b(\cdot)$ is accurately learned, then as sample size approaches infinity $\widehat\theta_{\text{debias}}$ will converge to the true value (i.e., the estimator is consistent) and is asymptotically normal under standard regularity conditions. Accurate learning of the bias implies it is not overfit; the fitting process should produce similar results if slightly different samples from the population of instances appeared in $D_{\text{shared}}$. 
Similar to the PPI estimator above, the precision of the corrected estimator $\widehat\theta_{\text{debias}}$ will depend on the size of $D_{\text{LLM}}$ relative to that of $D_{\text{shared}}$, as well as the variability of the target outcome and of the estimated bias. The key difference between this approach and the PPI approach above is that the former is only unbiased asymptotically. For a finite sample, even with many replications of the measurement process the expected value of the estimator will not necessarily equal the true value based in $\humandgp$.

\subsection{Examples of heuristic validation} %from \citet{anthis2025llm}}
\setlength{\tabcolsep}{4pt}
\renewcommand{\arraystretch}{1.15}

\begin{longtable}{|p{0.35\textwidth}|p{0.65\textwidth}|}
\hline
\textbf{Paper} & \textbf{Validation Description} \\
\hline
\endfirsthead

\hline
\textbf{Paper} & \textbf{Validation Description} \\
\hline
\endhead

\hline
\endfoot

\hline
\endlastfoot

Abdurahman, S., Atari, M., Karimi-Malekabadi, F., Xue, M., Trager, J., Park, P., Golazizian, P., Omrani, A., and Dehghani, M. (2024) Perils and Opportunities in Using Large Language Models in Psychological Research. \emph{PNAS Nexus}. &
GPT-3.5 administered MFQ-2 1{,}000 times; compared mean response magnitude, variance, and item-level correlations to human survey data; noted exaggerated cross-domain correlations and lower within-domain variance. Emphasis on replicating qualitative and quantitative effect patterns rather than formal hypothesis testing.\\
\hline
Abeliuk, A., Gaete, V., and Bro, N. (2025). ``Fairness in LLM-Generated Surveys.'' arXiv Preprint.& Benchmarked LLM-simulated survey responses against ground-truth demographic labels and annotated data. (i) individual level accuracy of predictions; (ii) distributional similarity between LLM- and human-derived responses using Jensen-Shannon similarity; (iii) subgroup parity tests (differences in mean response probability by demographic group); (v) comparison to a Random Forest trained on the same human data to contextualize predictive accuracy. \\
\hline
Aher, G., Arriaga, R. I., and Kalai, A. T. (2023). ``Using Large Language Models to Simulate Multiple Humans and Replicate Human Subject Studies.'' ICML.& Constructed parallel LLM and human experiments and compared summary results. (i) sign and direction of effects matched to original human studies; (ii) magnitude comparison via mean differences and standard errors on the same tasks; (iii) distributional comparisons to human results (graphical); (iv) evaluation on both replicated and unseen stimuli to test generalization beyond training exposure. Emphasis on replicating qualitative and quantitative effect patterns rather than formal hypothesis testing. Also checked that validity rate of LLM responses (proportion valid generations) is high and for consistency of responses for name pairs. \\
  \hline
  Ahnert, G., Pellert, M., Garcia, D., and Strohmaier, M. (2025). ``Extracting Affect Aggregates from Longitudinal Social Media Data with Temporal Adapters for Large Language Models.'' \textit{Proceedings of the International AAAI Conference on Web and Social Media}, 19(1), 15-36. https://doi.org/10.1609/icwsm.v19i1.35801 & Evaluated whether temporally fine-tuned LLMs reproduce population-level affect signals. (i) time-series correlation of LLM-derived affect aggregates (weekly) with YouGov survey affect indices, using Pearson's $r$ and permutation-based p-values; (ii) examination of mean and median squared errors and standard deviation of squared error computed against human data; (iii) trend alignment across major events (graphical qualitative analysis); (iv) comparison against BERT baseline models and statistical nulls.\\
  \hline
  Argyle, L., Busby, E., Fulda, N., Gubler, J., Rytting, C. and Wingate. D. (2023). ``Out of One, Many: Using Language Models to Simulate Human Samples.'' Political Analysis.& Proposed ``algorithmic fidelity'' as a construct for LLM-human comparison. (i) human-judge evaluations of free-text responses for human-likeness, including statistical test of difference in proportion of human lists that people judged to be human versus proportion of GPT generated lists they judged to be human and comparisons of prevelance of traits humans rated in human versus GPT lists (fit multiariate regression to traits and whether it was a human or GPT list, compare coefficients) ; (ii) correlation between probability of voting for human and GPT responses on ANES data, (iii) Cramer’s $V$ and correlation matrices across categorical and ordinal variables capturing voting and demographic and political preference variables between LLM-generated and ANES survey samples. Integrated both quantitative similarity metrics and qualitative human judgments of realism.\\
  \hline
  Binz, M., and Schulz, E. (2023). ``Turning Large Language Models into Cognitive Models.'' ICLR.& Fine-tuned LLaMA embeddings with a logistic layer on human trial-level data and validated by: (i) predictive accuracy (log-likelihood) on held-out human trials from Choices13k and horizon tasks; (ii) compare goodness of fit for random model, LLM without finetuning, domain specific model (iii) graphical comparison of patterns in model predictions relative to human data (iv) proportion of participants whose individual level data was best explained by Centaur model (v) cross-task generalization to an unseen task. Emphasis on replicating quantitative fit patterns and out-of-sample predictive accuracy rather than hypothesis testing.\\
  \hline
  Binz, M., Akata, E., Bethge, M., Brändle, F., Callaway, F., Coda-Forno, J., Dayan, P., Demircan, C., Eckstein, M.K., Éltető, N., and Griffiths, T.L. (2025). ``Centaur: A Foundation Model of Human Cognition.'' arXiv Preprint.& QLoRA-finetuned Llama on Psych-101 dataset of 160 behavioral experiments. (i) predictive accuracy (negative log likelihood) on held-out participants and experiments; (ii) out-of-domain generalization across domains and cover-story changes; (iii) predictive accuracy compared to domain-specific cognitive models (iv) qualitative pattern checking on open-loop simulation results (check match to human behavioral patterns when model generates new human behavior when fed its own output) (v) variance explained in human reaction-time data via mixed-effects models; (v) predictive accuracy of model activations for human fMRI patterns.\\
  \hline
  Bisbee J, Clinton JD, Dorff C, Kenkel B, and Larson JM. (2024). Synthetic Replacements for Human Survey Data? The Perils of Large Language Models. Political Analysis.& Compared ChatGPT personas to ANES respondents. (i) comparison of means and variances for feeling-thermometer ratings; (ii) regression coefficient comparison for vote choice and issue positions; (iii) identification of sign flips and magnitude discrepancies (48\% significant differences, 32\% sign reversals); (iv) prompt and model version variation tests across time.\\
  \hline
  Boelaert, J., Coavoux, S., Ollion, É., Petev, I., and Präg, P. (2025). Machine Bias. How Do Generative Language Models Answer Opinion Polls?. Sociological Methods \& Research.& Prompted LLMs with survey questions modeled after World Values Survey (WVS) items and compared distributional statistics to human data from multiple waves of WVS across five countries. (i) comparison of response means and variances to human poll responses; (ii) examination of variance attenuation and systematic directional bias (``machine bias''); (iii) assessment of topic-level instability across prompts. For direct comparisons of LLM-predicted output with human data, emphasis on qualitative and descriptive pattern comparison using a normalized "Earth mover distance" (nEMD) metric. Use statistical models to estimate whether demographic attributes predict nEMD (evaluating for bias) and BIC-based goodness of fit comparisons to evaluate whether bias is better explained by machine-error or social/demographic factors. \\
  \hline
  Brand, J., Israeli, A., Ngwe, D. (2023). ``Using GPT for Market Research.'' ACM Conference on Economics and Computation.& Queried GPT to simulate consumer choice behavior. (i) economic rationality checks (downward-sloping demand, income effects, state dependence); (ii) comparison of willingness-to-pay magnitudes to human conjoint study benchmarks; (iii) fit of multinomial logit and regression models to LLM-generated data and comparison of coefficients and signs to human estimates; (iv) reported standard errors for synthetic samples. Emphasis on approximate quantitative pattern replication. \\
  \hline
  Chen, X., Kirshner, S.N., Ovchinnikov, A., Andiappan, M. and Jenkin, T. (2025). ``A Manager and an AI Walk into a Bar: Does ChatGPT Make Biased Decisions Like We Do.'' Manufacturing and Service Operations Management.& Designed decision-making tasks illustrating known human cognitive biases (e.g., anchoring, loss aversion, framing). (i) compared bias directionality and strength (significance) to benchmark human experiments; (ii) comparison of bias to theoretical rationality expectations; (iii) assessed stability across replications.\\
  \hline
  Dominguez-Olmedo, R., Hardt, M. and Mendler-Dünner, C. (2024). Questioning the survey responses of large language models. NeurIPS.& Used LLMs to simulate American Community Survey responses. Compares (numerically and visually) entropy, KL divergence, and systematic bias towards the answer choice "A" for simulated responses against the reference distribution produced by the 2019 ACS data. \\
  \hline
  Gao, Y., Lee, D., Burtch, G., and Fazelpour, S. (2024). ``Take Caution in Using LLMs as Human Surrogates.'' arXiv Preprint.& Tested LLM behavior in a modified Keynesian beauty contest (11-20 money request game). (i) compared distribution of strategic depth to human experiment data; (ii) evaluated effect of prompt language variation and rationale provision; (iv) examined fine-tuning and retrieval-augmentation methods. Statistical testing included permutation tests of Jensen-Shannon divergence of human vs. LLM distributions of responses. Emphasis on descriptive comparisons and qualitative pattern replication.\\
  \hline
  Gerosa, M., Trinkenreich, B., Steinmacher, I. and Sarma, A. (2024). ``Can AI serve as a substitute for human subjects in software engineering research?.'' Automated Software Engineering.& Simulated open-source contributor survey responses, focus groups, and interviews. (i) compared item-level Likert response means and proportions between LLM- and human surveys; (ii) qualitatively inspected free-text explanations for face validity.\\
  \hline
  Gonzalez-Bonorino, A., Capra, C.M., and Pantoja, E. (2025). ``LLMs Model Non-WEIRD Populations: Experiments with Synthetic Cultural Agents.'' arXiv Preprint.& Modeled Dictator Game, Ultimatum Game, and Endowment Effect tasks across simulated small-scale societies. (i) compared mean offers and acceptance rates to ethnographic benchmarks; (ii) tested for simulated cross-cultural differences. Statistical testing (Cochran-Mantel-Haenszel test of variations in acceptance decisions across offer levels conditional on tribal affiliation) to evaluate whether simulated agents exhibited variation across (simulated) cultural groups. Qualitative comparison of simulations against human sample benchmarks.\\
  \hline
  Gui, G., and Toubia, O. (2023). ``The Challenge of Using LLMs to Simulate Human Behavior: A Causal Inference Perspective.'' arXiv Preprint.& Used LLM personas to simulate consumer choice and willingness-to-pay studies. (i) compared aggregate WTP means and price elasticities to human survey data; (ii) tested prompt variants with demographic attributes; (iii) evaluated simulation of individual responses ("blinded" to the experimental design) against prompting to directly elicit experimental treatment effects estimates ("unblinded"). Emphasis on descriptive patterns, visualizations, and qualitative alignment to economic theory.\\
  \hline
  Hewitt, L., Ashokkumar, A., Ghezae, I., and Willer, R. (2024). ``Predicting Results of Social Science Experiments Using Large Language Models.'' Self-hosted Preprint.& Simulated 70 preregistered survey experiments from TESS. (i) computed correlation of GPT-4 predicted treatment effects with human average treatment effects \(r \approx 0.85\text{--}0.94\); (ii) compared aggregate effect magnitudes and signs; (iii) examined out-of-distribution performance on unpublished studies; (iv) analyzed error patterns by study type.\\
  \hline
  Heyman, T., and Heyman, G. (2024). ``The impact of ChatGPT on human data collection: A case study involving typicality norming data.'' Behavior Research Methods.& Replicated typicality rating studies for category memberships. (i) correlation of LLM ratings with mean human typicality scores; (ii) item-total correlations per category; (iii) test-retest correlations of LLM ratings within categories.\\
  \hline
  Horton, J. (2023). ``Large Language Models as Simulated Economic Agents: What Can We Learn from Homo Silicus?.'' NBER Working Paper.& Replicated canonical behavioral economics games using GPT agents. (i) compared mean offer rates, acceptance probabilities, and bid distributions to human lab data across ultimatum, dictator, and public goods games; (ii) checked directional consistency of treatment effects (e.g., punishment or framing manipulations); (iii) verified monotonicity of responses to payoff structure changes; (iv) qualitatively inspected strategy rationales.\\
  \hline
  Hämäläinen, P., Tavast, M. and Kunnari, A. (2023). ``Evaluating Large Language Models in Generating Synthetic HCI Research Data: a Case Study.'' ACM CHI.& Analyzed text responses describing art experiences in video games from a prior study and prompted comparable simulated data using LLMs. (i) collected human evaluators' ratings of whether texts were generated by humans or AIs (randomly assigned pairwise comparisons of texts drawn from each set); (ii) human evaluators' reasons for their decisions in i; (iii) qualitative analysis and keyword frequency comparisons of human evaluators' open-ended descriptions of the texts generated as part of i; and (iv) visual and descriptive quantitative analysis of semantic clustering overlap (using LLM to code/cluster/reduce texts generated by people/LLMs). Statistical testing in (i) through paired t-tests for differences of proportions (framed as borrowing from signal detection theory methods). \\
  \hline
  Jiang, H., Zhang, X., Cao, X., Breazeal, C., Roy, D., and Kabbara, J. (2024). ``PersonaLLM: Investigating the Ability of Large Language Models to Express Personality Traits.'' NAACL.& Modeled Big Five personality questionnaires in GPT-based agents. (i) compared mean factor scores (OCEAN dimensions) between prompts for different personality types (paired T tests); evaluated free text responses between humans and LLM-personas in terms of (ii) word choices (LIWC); and human ratings vs machine ratings in terms of pairwise correlations. Emphasis on correlation and qualitative trend matching. \\
  \hline
  Jiang, S., Wei, L., and Zhang, C. (2025). ``Donald Trumps in the Virtual Polls: Simulating and Predicting Public Opinions in Surveys Using Large Language Models.'' arXiv Preprint.& Used World Values Survey and ANES items to simulate cultural and political attitudes at an individual respondent level and to evaluate LLM simulation accuracy at aggregate (sample, population) and individual levels. (i) matched country-level means and standard deviations between LLM and survey data; (ii) computed individual-level correlations for ideological and trust constructs; (iii) inspected electoral college forecast outcomes from original survey data vs. LLM vs. human vote outcomes. Statistical tests include paired T-tests, correlation. Emphasis on qualitative trend matching, predictive accuracy (qualitative comparison of electoral forecasting in and out-of sample).\\
  \hline
  Kim, J., and Lee, D. (2024). ``AI-Augmented Surveys: Leveraging Large Language Models and Surveys for Opinion Prediction.'' arXiv Preprint. & Tested LLMs for recovering missing demographic and attitudinal variables. (i) computed prediction accuracy against held-out human responses; (ii) analyzed calibration and bias by subgroup; (iii) compared imputation distributions to ground truth marginals; (iv) tested fine-tuning with human-subject embeddings. Quantitative evaluation emphasizing predictive validity (AUC), correlation statistics.\\
  \hline
  Kozlowski, A.C., Kwon, H. and Evans, J.A. (2024). ``In Silico Sociology: Forecasting COVID-19 Polarization with Large Language Models.'' arXiv Preprint.& Replicated pandemic-related behavior and attitude surveys. (i) compared distributions of behavioral-intention responses between human and LLM samples; (ii) measured correlation between simulated and real treatment effects (e.g., message framing); (iii) examined subgroup means across demographics; (iv) inspected textual rationales for interpretability.\\
  \hline
  Lampinen, A.K., Dasgupta, I., Chan, S.C., Sheahan, H.R., Creswell, A., Kumaran, D., McClelland, J.L. and Hill, F. (2024). ``Language models, like humans, show content effects on reasoning tasks.'' PNAS Nexus.& Tested LLMs on classic logical reasoning tasks (e.g., Wason selection task). (i) compared proportions of logically correct responses to human benchmark data; (ii) examined qualitative explanation similarity for correct vs. incorrect reasoning; (iii) computed variance in correct-answer frequency across prompts; (iv) visual inspection of reasoning rationales.\\
  \hline
  Lin, J., Sun, L. and Yan, Y., et al. (2025). ``Simulating Macroeconomic Expectations using LLM Agents.'' arXiv Preprint.& Used Michigan Survey of Consumers as a benchmark. (i) compared mean consumer sentiment indices across time between LLM- and survey-based samples; (ii) computed Pearson correlations of monthly aggregates; (iii) compared trend alignment and turning points; (iv) analyzed sensitivity to demographic prompt variation. \\
    \hline
  Liu, R., Geng, J., Peterson, J.C., Sucholutsky, I. and Griffiths, T.L. (2024a). ``Large language models assume people are more rational than we really are.'' arXiv Preprint.& Simulated human choices from the Choices13k dataset using CoT prompting. (i) compared accuracy of next-choice predictions to human-level baselines; (ii) measured calibration of predicted choice probabilities; (iii) evaluated changes in predictive accuracy when adding human reasoning examples; (iv) qualitative inspection of generated rationales.\\
    \hline
  Lyman, A., Hepner, B., Argyle, L. P., Busby, E. C., Gubler, J. R., and Wingate, D. (2025). ``Balancing Large Language Model Alignment and Algorithmic Fidelity in Social Science Research.'' Sociological Methods \& Research.& Compared multiple alignment-tuned vs. base LLMs on political attitude simulations. (i) correlation of LLM-simulated partisan attitude distributions with ANES human data; (ii) analysis of ``algorithmic fidelity'' assessing whether completions adhere to logical relations posited within prompts; (iii) comparison of key attitude–behavior relationships across models; (iv) LLM and human judge coding of responses. Quantitative and qualitative evaluation of alignment vs fidelity, statistical comparisons and uncertainty calculated using permutation methods/tests.\\
    \hline
  Manning, B.S., Zhu, K., and Horton, J. J. (2024). ``Automated Social Science: Language Models as Scientist and Subjects.'' arXiv Preprint.& Used LLMs to both generate experimental hypotheses and simulate participants in auction and bargaining tasks. (i) compared mean treatment effects from LLM-simulated auctions to human lab results; (ii) checked sign consistency of economic effects (e.g., reserve price, information symmetry); (iii) correlated estimated coefficients from LLM and human regressions; (iv) examined LLM generated hypotheses for plausibility relative to published literature. \\
    \hline
  Meister, N., Guestrin, C., and Hashimoto, T. (2025). ``Benchmarking Distributional Alignment of Large Language Models.'' NAACL.& (i) evaluate distributional alignment to human opinion distributions under three distribution expression methods (model log-probs, sampling, verbalized proportions); (ii) compute distances to human group distributions including Total Variation and related metrics; (iii) compare steering methods and question domains; (iv) include human baselines and rank models vs humans.\\
    \hline
  Moon, S., Abdulhai, M., Kang, M., Suh, J., Soedarmadji, W., Behar, E.K. and Chan, D.M. (2024). ``Virtual Personas for Language Models via an Anthology of Backstories.'' EMNLP.& Administer three Pew American Trends Panel surveys to LLM personas; (i) measure distributional similarity to human response distributions using Wasserstein distance; (ii) assess response consistency metrics; (iii) report percentage improvement over baselines for under-represented subgroups.\\
    \hline
  Park, J.S., Zou, C., Shaw, A.,Hill, B.M., Cai, C., Morris, M.R., Willer, R., Liang, P., and Bernstein, M. (2024). ``Generative Agent Simulations of 1,000 People.'' arXiv Preprint.& Construct interview-style backstories then run widely-used survey and behavioral tasks. (i) compare sign/direction and magnitude of effects to human studies and public datasets; (ii) graphical and distributional comparisons across tasks; (iii) check internal response-quality/consistency of generated personas. Statistical comparisons (including hypothesis tests against null of no differences between different agent types) of correlations, MAE between several agent approaches and human participants (self-consistency).\\
    \hline
  Park, J. S., O'Brien, J., Cai, C. J., Morris, M. R., Liang, P., Bernstein, M. S. (2023). ``Generative Agents: Interactive Simulacra of Human Behavior.'' UIST.& (i) Controlled evaluation by ``interviewing'' agents to test staying in character, memory, planning, reaction, and reflection; (ii) End-to-end town simulation over two in-game days showing emergent social behaviors (e.g., invitations disseminated and coordinated attendance at a Valentine’s Day party); (iii) Ablations of memory retrieval, planning, and reflection to establish each component’s contribution to believability; (iv) qualitative error analysis of failures such as missed memory retrieval and over-formal speech. Quantitative evaluation focuses on ELO-type (TrueSkill) scores obtained from human raters evaluating believability of interactions transcript produced by full generative agent architecture vs. ablated conditions and a control transcript produced by human crowdworkers observing agent interactions. Statistical testing using Kruskall-Wallis test on raw rank data for believability evaluations, Dunn post-hoc test to compare pairwise differences.\\
    \hline
  Pellert, M., Lechner, C.M., Wagner, C., Rammstedt, B. and Strohmaier, M. (2024). ``AI Psychometrics: Assessing the Psychological Profiles of Large Language Models Through Psychometric Inventories.'' Perspectives on Psychological Science.& Administer MFQ items to LLMs; (i) compare mean domain scores and response distributions to human MFQ norms; (ii) inspect cross-item patterns and variance for human-likeness. Emphasis on descriptive alignment checks and pattern diagnostics.\\
  \hline
  Petrov, N.B., Serapio-García, G. and Rentfrow, J. (2024). Limited ability of llms to simulate human psychological behaviours: a psychometric analysis. arXiv Preprint.& Elicit Big Five, PANAS, BPAQ, SSCS from persona-steered LLMs; (i) compare item- and scale-level distributions to human benchmarks; (ii) report item-level correlations and shifts relative to human norms; (iii) analyze how persona prompts move trait profiles.\\
  \hline
  Qu, Y., and Wang, J. (2024). ``Performance and biases of large language models in public opinion simulation.'' Humanities \& Social Sciences Communications.& Simulate WVS responses under demographic conditioning; (i) quantify distributional similarity and prediction error vs human country- and subgroup-level distributions; (ii) subgroup bias analyses across themes and regions; (iii) ablations on prompt settings.\\
  \hline
  del Rio-Chanona, R.M., Pangallo, M. and Hommes, C.. (2025). Can Generative AI agents behave like humans? Evidence from laboratory market experiments. arXiv preprint.& Run market-like mechanisms with LLM agents. (i) compare resulting price signals to human or realized outcomes; (ii) evaluate forecast accuracy with error metrics and correlations; (iii) benchmark against simple baselines and alternative agent configurations.\\
  \hline
  Ross, J., Kim, Y., and Lo, A. W. (2024). ``LLM economicus? Mapping the Behavioral Biases of LLMs via Utility Theory.'' COLM.& Estimate utility-theory parameters from LLM choices in ultimatum, loss-aversion, risk, and time-discounting tasks. (i) compare fitted parameters to human estimates from the literature; (ii) analyze prompt-role effects (role-play vs advice) on parameter shifts; (iii) report goodness-of-fit and parameter confidence.\\
  \hline
  Santurkar, S., Durmus, E., Ladhak, F., Lee, C., Liang, P. and Hashimoto, T.. (2023). ``Whose Opinions Do Language Models Reflect?'' ICML.& Build OpinionsQA from Pew ATP. (i) compute alignment of model response distributions with 60 demographic groups as an average 1-Wasserstein–based similarity between the LM’s distribution and human group distributions:; (ii) regression analyses of demographic misalignment and steering effects; (iii) compare HF-tuned vs base models.\\
  \hline
  Suh, J., Jahanparast, E., Moon, S., Kang, M., and Chang, S. (2025). ``Language Model Fine-Tuning on Scaled Survey Data for Predicting Distributions of Public Opinions.'' ACL.& Fine-tuned LLMs on SubPOP to match human group-level survey distributions. (i) 1-Wasserstein distance between LLM option distributions (from token log-probs) and human distributions; (ii) KL divergence as auxiliary diagnostic; (iii) human bootstrap lower bound as a ceiling; (iv) generalization tests to unseen waves, groups, and a different survey (ATP → GSS) with 38-54\% WD gap reduction vs prompting baselines.\\
  \hline
  Taubenfeld, A., Dover, Y., Reichart, R., and Goldstein, A. (2024). ``Systematic Biases in LLM Simulations of Debates.'' EMNLP.& Simulated partisan debates with LLM agents and compared attitude-change dynamics to established human patterns from the literature. (i) tracked within-debate agreement shifts and sign consistency vs known human effects; (ii) analyzed convergence toward model-intrinsic biases even when role-playing opposing sides; (iii) ran self-fine-tuning interventions to causally alter agent bias and observed behavioral shifts; (iv) qualitative comparison to human social dynamics findings.\\
  \hline
  Toubia, O., Gui, G.Z., Peng, T., Merlau, D.J., Li, A. and Chen, H. (2025). ``Database Report: Twin-2K-500: A Data Set for Building Digital Twins of over 2,000 People Based on Their Answers to over 500 Questions.'' Marketing Science.& Dataset with 4 survey waves (N=2,058; 500 questions). (i) establish a human test-retest ceiling by repeating tasks in Wave 4; (ii) train twins on Waves 1-3, then evaluate individual-level predictive accuracy on held-out items relative to the test–retest ceiling and a random benchmark; (iii) aggregate-level checks: test whether twins replicate average treatment effects for 11 between-subject and 5 within-subject heuristics-and-biases studies, reporting where replication succeeds or fails; (iv) pricing study sanity checks: compare demand curves from Wave 3 vs Wave 4 vs twins and inspect expected properties (downward slope for positive prices).\\
  \hline
  Tranchero, M., Brenninkmeijer, C.F., Murugan, A. and Nagaraj, A. (2024). ``Theorizing with Large Language Models.'' NBER Working Paper.& Used LLM agents to replicate a human exploration-exploitation lab experiment and then extend it. (i) aggregate outcome matching to human subjects on the original task via side-by-side plots of average group earnings and breakthrough probability across information conditions; (ii) directional effect checks for herding vs coordinated search under varying payoff information and rivalry; (iii) mechanism/boundary tests by varying environment parameters and agent objectives, assessing whether qualitative patterns track human theory and results; (iv) baseline sanity test comparing direct GPT-4 outcome predictions to both human and agent-based results, showing weaker alignment for direct prediction than for agent simulations. \\
  \hline
  Wang, A., Morgenstern, J. and Dickerson, J.P. (2025). ``Large language models that replace human participants can harmfully misportray and flatten identity groups.'' Nature Machine Intelligence.& Ran human studies with 3,200 participants across 16 identities and compared to identity-prompted LLM outputs. (i) Misportrayal: measured distances between LLM responses and in-group vs out-group human responses using ngram, SBERT, and Multiple Choice Wasserstain distance, both as average and closest-match metrics; significance via two-sided Welch’s t-tests with CIs/bootstraps. (ii) Flattening: tested loss of within-group diversity via SBERT covariance trace, SBERT pairwise distance, n-gram uniqueness, and MC uniques. (iii) Also analyzed coverage diversity using SBERT covariance determinant and Vendi score; reported inference-time mitigations (identity-coded names, temperature). Statistical testing and confidence intervals reported – Quantitative evaluation of group-level divergence and variance structure.\\
  \hline
  Wang, P., Zou, H., Yan, Z., Guo, F., Sun, T., Xiao, Z., and Zhang, B. (2024). ``Not Yet: Large Language Models Cannot Replace Human Respondents for Psychometric Research.'' OSF Preprint.& Evaluated LLM-generated responses on BFI-2 and HEXACO-100 against human data. (i) compared descriptive statistics at item, facet, and domain levels (means, SDs) for LLM vs human samples; (ii) conducted psychometric modeling with model fit indices and factor loadings to test whether LLM responses recover the intended latent structures; (iii) compared inter-factor correlation matrices between LLM and human data; (iv) analyzed multiple LLM response modes and prompt variants to test robustness. \\
  \hline
  Weidmann, B., Xu, Y. and Deming, D.J. (2025). ``Measuring Human Leadership Skills with AI Agents.'' NBER Working Paper.& Pre-registered lab study that (i) builds an AI-leadership test by assigning human leaders to teams of AI agents and measuring group performance; (ii) independently estimate each leader's causal impact on human teams by repeatedly randomizing leaders to human groups; (iii) report correlation (0.81) between AI-team leadership and human-team leadership; (iv) analyze behavioral predictors (question asking, turn-taking) and cognitive covariates. \\
  \hline
  Xie, C., Chen, C., Jia, F., Ye, Z., Lai, S., Shu, K., Gu, J., Bibi, A., Hu, Z., Jurgens, D. and Evans, J. (2024). ``Can Large Language Model Agents Simulate Human Trust Behavior?'' NeurIPS.& Trust behavior simulations across Trust Game variants. (i) compare amount sent distributions of LLM agents vs human studies; include Valid Response Rate (VRR) for budget compliance and medians by model size; (ii) test behavioral factors via contrasts that mirror human findings: Trust vs Dictator (reciprocity anticipation), MAP Trust vs Risky Dictator (risk perception), and prosocial-preference signals; (iii) analyze behavioral dynamics in repeated Trust Game and plot trust-rate curves; (iv) probe agent vs human target and demographic persona effects; (v) inspect BDI/CoT reasoning consistency with actions.\\
  \hline
  Zhang, S., Xu, J. and Alvero, A.J.. (2025). ``Generative AI Meets Open-Ended Survey Responses: Research Participant Use of AI and Homogenization.'' Sociological Methods \& Research.& Human to LLM text comparison using open-ended prompts from pre-ChatGPT studies. (i) evaluate homogenization via higher average pairwise similarity and reduced lexical diversity, and positivity shift via sentiment measures; (iii) sensitivity on social group descriptions, showing larger homogenization and positivity for sensitive topics; (iv) robustness across models and prompt variants.\\
  \hline
  Zhang, X., Lin, J., Mou, X., Yang, S., Liu, X., Sun, L., Lyu, H., Yang, Y., Qi, W., Chen, Y. and Li, G. (2025). ``SocioVerse: A World Model for Social Simulation Powered by LLM Agents and A Pool of 10 Million Real-World Users.'' arXiv Preprint& Large-scale world-model with four alignment modules and a 10M user pool. (i) Politics: simulate US presidential election dynamics, compare state-level accuracy rate and RMSE to polling/ground truth; (ii) News/breaking news feedback loops: normalized RMSE measures point-wise differences between simulations and ground truth, and KL-divergence compares the distributions to observed user-pool statistics; (iii) Economics: simulate China national economic surveys, assess representativeness and distributional similarity of macro attitudes vs human benchmarks, again using NRMSE and KL-divergence; (iv) ablations for each alignment module and agent–user-pool matching checks.\\
  \end{longtable}

\begin{comment}
\onecolumn
\appendix

\iftoggle{icml}{\onecolumn}{}
\thispagestyle{empty}
{%
\begin{center}
{\bfseries Supplementary Materials}    
\end{center}
}
\pagenumbering{arabic}
\renewcommand*{\thepage}{\arabic{page}}

\section{Notation}
\label{Appendix::Notation}

We provide a list of the notation used throughout the paper in~\cref{Table::Notation}
\end{comment}

\end{document}